%% file: main.tex
\documentclass{article}

\usepackage{arxiv}
\usepackage{verbatim}
\usepackage[T1]{fontenc}    
\usepackage{hyperref}       
\usepackage{url}            
\usepackage{booktabs}       
\usepackage{amsfonts}       
\usepackage{nicefrac}       
\usepackage{microtype}      
\usepackage{graphicx}
\usepackage{natbib}
\usepackage{doi}
\usepackage{multirow}
\usepackage{amsmath}

\newcommand{\tabincell}[2]{\begin{tabular}{@{}#1@{}}#2\end{tabular}}
\newcommand{\modelname}{\textsc{AliceMind-MMU }}
\usepackage{subfigure}
\usepackage[utf8x]{inputenc}
\usepackage{xcolor}

\title{Achieving Human Parity on Visual Question Answering}

\author{
Ming Yan\thanks{Equal contribution}\hspace{2mm}, Haiyang Xu$^*$, Chenliang Li$^*$, Junfeng Tian$^*$, Bin Bi$^*$, Wei Wang, Weihua Chen, Xianzhe Xu,
\\ \textbf{Fan Wang, Zheng Cao, Zhicheng Zhang, Qiyu Zhang, Ji Zhang, Songfang Huang, Fei Huang, Luo Si, Rong Jin}
\\Alibaba Group\\
    \{ym119608, shuofeng.xhy, lcl193798, tjf141457, b.bi, hebian.ww, kugang.cwh, xianzhe.xxz, \\
     fan.w, zhengzhi.cz, zhangzhicheng.zzc, qiyu.zhang, zj122146, songfang.hsf, f.huang, luo.si, jinrong.jr\}@alibaba-inc.com
}
\date{}

\renewcommand{\undertitle}{}

\begin{document}
\maketitle

\begin{abstract}
The Visual Question Answering (VQA) task utilizes both visual image and language analysis to answer a textual question with respect to an image. It has been a popular research topic with an increasing number of real-world applications in the last decade. This paper describes our recent research of AliceMind-MMU (ALIbaba's Collection of Encoder-decoders from Machine IntelligeNce lab of Damo academy - MultiMedia Understanding) that obtains similar or even slightly better results than human being does on VQA. This is achieved by systematically improving the VQA pipeline including: (1) pre-training with comprehensive visual and textual feature representation; (2) effective cross-modal interaction with learning to attend; and (3) A novel knowledge mining framework with specialized expert modules for the complex VQA task. Treating different types of visual questions with corresponding expertise needed plays an important role in boosting the performance of our VQA architecture up to the human level. An extensive set of experiments and analysis are conducted to demonstrate the effectiveness of the new research work.
\end{abstract}


\section{Introduction}
\input{intro}

\section{VQA Modeling}
\label{sec:model}
\subsection{Reaching Human Parity}
\input{human_parity}

\begin{figure} 
    \centering
    \includegraphics[width=0.8\textwidth]{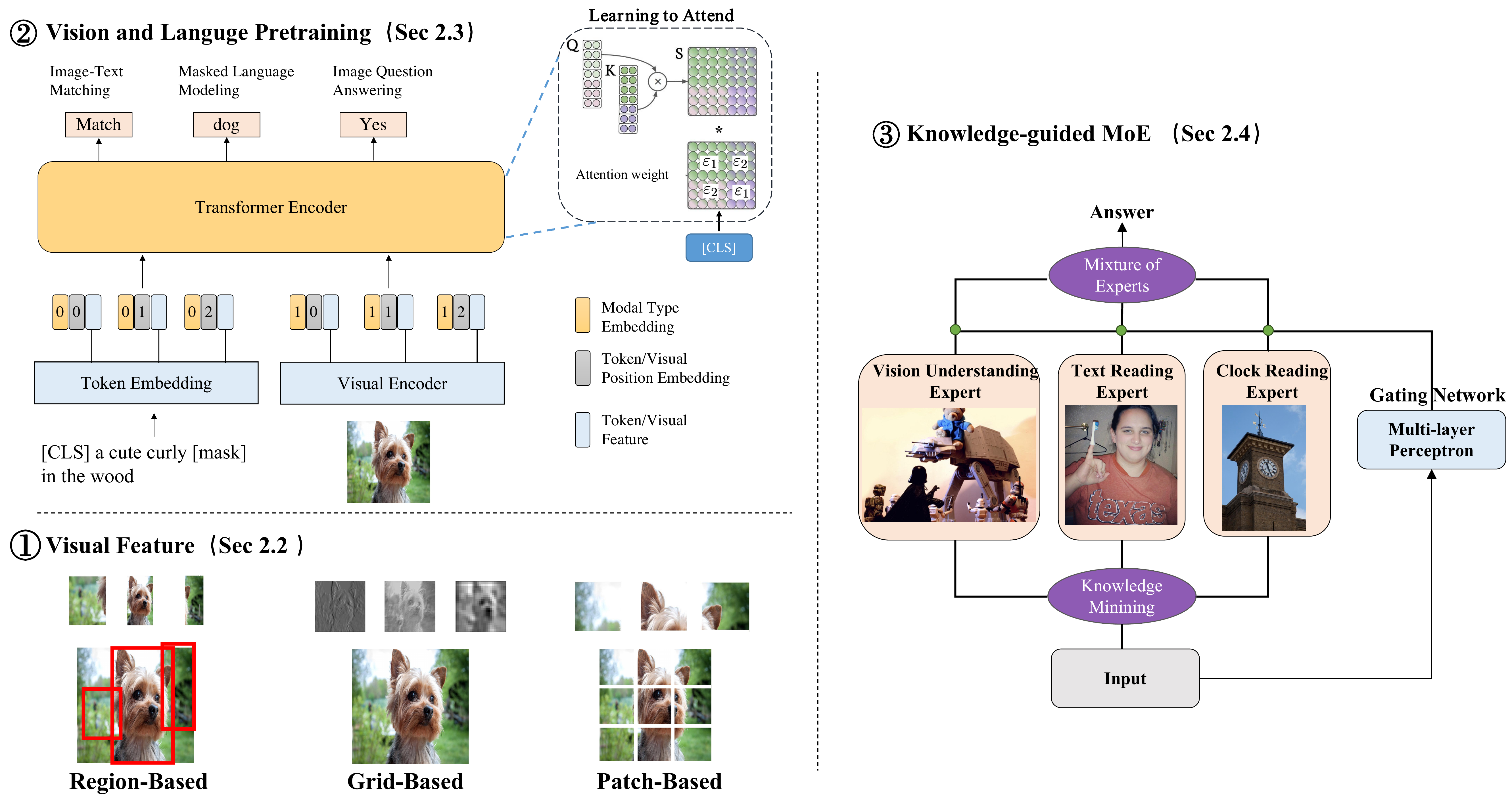}
    \caption{The framework of \modelname.}
    \label{framework}
\end{figure}

\subsection{Comprehensive Feature Representation}

Feature representation for vision and text is fundamental for cross-modal learning. Different kinds of features can help capture diverse data characteristics, which complements with each other.

\subsubsection{Visual Features} \label{visual_fea}
For comprehensive visual feature understanding, three kinds of visual features are considered: \textit{region feature}, \textit{grid feature} and \textit{patch feature}. 

\paragraph{Region Feature} 
With the discovery of `bottom-up' attention~\citep{anderson2018bottom}, region-based visual features have been the \textit{de facto} standard for vision and language tasks. Unlike normal `top-down' attention that directly focuses on semantically irrelevant parts of visual input, bottom-up attention uses pre-trained object detectors~\citep{ren2015faster} to identify salient regions based on the visual input. As a result, images are represented by a collection of region-based features, which provide better localization of individual objects and capture the detailed semantics within the image content. Generally, region-based visual encoders such as BUTD~\citep{anderson2018bottom} are pre-trained with detection data like Visual Genome~\citep{krishna2017visual}. Recently, VinVL~\citep{zhang2021vinvl} has been built on a large-scale pre-trained object-attribute detection model with much larger amounts of data on four public object detection datasets, which helps better capture both coarse-level and fine-grained visual semantic information in images. The object detector from VinVL~\citep{zhang2021vinvl} is used in this work to extract a collection of object-level region features with more detailed visual semantics, where each object $o_j$ is represented as a 2048-dimensional feature vector $r_j$. To capture the spatial information of the object, box-level location features for each object are also encoded via a 4-dimensional vector $l_j=(\frac{x_1}{W},\frac{y_1}{H},\frac{x_2}{W},\frac{y_2}{H})$ as in SemVLP~\citep{li2021semvlp} and LXMERT~\citep{tan2019lxmert}. The $r_j$ and $l_j$ are concatenated to form a position-sensitive object feature vector, which is further transformed to a lower dimension of $D$ using a linear projection to ensure that it has the same vector dimension as that of token embeddings.

Despite the superior performance obtained via region-based visual features, this kind of features suffer from several problems. Firstly, the region-based methods heavily rely on a pre-trained object detector, where the performance may be bounded by the capability of the object detector and its predefined visual vocabulary. Besides, only salient regions of image are used in region-based methods, where the global or background information may be missing. 

\paragraph{Grid Feature}
To address the limitations of region-based features like locality, some work such as PixelBERT~\citep{huang2020pixel}, E2E-VLP~\citep{xu2021e2e}, Grid-VLP~\citep{yan2021gridvlp}
and \citep{jiang2020defense} have been proposed to revisit grid-based convolutional features for multi-modal learning, skipping the expensive region-related steps. The advantage lies in that: 1) the grid-based feature allows to introduce more flexible architectural designs for vision and language tasks, which makes it possible to support end-to-end training and efficient online inference; 2) it operates on a full image instead of a collection of semantic regions, so it can better capture global information of an image such as the background information; 3) it does not rely on a pre-trained object detector with limited visual vocabulary. Specifically, starting from the raw image $I_{img} \in R^{3\times H_0\times W_0}$ with 3 color channels, a fixed CNN-based image encoder such as ResNet~\citep{he2016deep} generates a lower-resolution activation map $F_{img} \in R^{C\times H \times W}$, where $C$ is the channel width and $H=\frac{H_0}{32}$, $W=\frac{W_0}{32}$. As the cross-modal fusion network expects a sequence as input, the spatial dimensions of $F_{img}$ are collapsed into one dimension, resulting in a $HW\times C$ feature map. Finally, a linear projection layer is used to reduce the channel dimension of the high-level feature map from $C$ to a smaller dimension $D$ for matching the dimension of token embeddings. To distinguish between different modalities, the grid feature map is supplemented with a learnable modal type embedding which is added to the output of linear projection layer.

To generate good grid features, it is very important to pre-train a strong CNN-based image encoder, to which the visual semantic information is incorporated. In terms of the data used to pre-train the image encoder, there are mainly two ways along this line: 1) \textit{Supervised Pre-training}: the image encoder is pre-trained with image classification data such as ImageNet~\citep{deng2009imagenet} or detection data such as Visual Genome~\citep{krishna2017visual}. As found in~\citep{jiang2020defense}, the large-scale object and attribute annotations collected in the Visual Genome are very helpful to provide the grid feature with visual semantics incorporated; 2) \textit{Unsupervised Pre-training}: the image encoder is pre-trained with a large amount of unlabeled image-text pairs without human supervision such as CLIP~\citep{radford2021learning}, where about 400M aligned image-text pairs are used. It belongs to the line of research that learns visual representations from natural language supervision~\citep{jia2021scaling,desai2021virtex}. In this way, the image encoder is naturally aligned with the textual semantics to facilitate the cross-modal fusion. It is well recognized that fully supervised pre-trained CNN model shows promising performance on in-domain or near-domain datasets, while it may not yield best performance when coming to transfer learning on out-domain datasets. Features derived from different ways can well complement with each other, which adapts to different kinds of questions.

\paragraph{Patch Feature}
Vision Transformer (ViT)~\citep{dosovitskiy2020image} has achieved outstanding performance in various visual tasks~\citep{liu2021swin, touvron2021training, chen2021pre}. It firstly splits an image into fixed-size patches, then uses a simple linear projection of a patch before feeding them into transformers. ViLT~\citep{kim2021vilt} is the first to explore patch-based features for multi-modal learning, and achieves up to dozens of times faster inference than previous region-based VLP methods. The advantages of patch-based features are following: 1) its simple framework can be more efficient than grid-based convolutional features in the online inference phrase; 2) it is more effective in capturing the global structure of a full image with the self-attention mechanism, which can provide complementary visual features different from region-based and grid-based ones. Specifically, the 2D image  $I_{img} \in R^{3\times H_0\times W_0}$ is reshaped into a sequence of flattened 2D patches $x_{p} \in R^{N\times (P^2 \cdot C)}$, where $(H_0, W_0)$ is the resolution of the original image, $C$ is the number of channels,$(P, P)$  is the resolution of each image patch, and $N = HW/P^2$ is the resulting number of patches and also serves as the input sequence length for the Transformer. Then, the patches are flatten and embedded to $D$ dimensions with a trainable linear projectionm, and an appropriate position encoding is introduced to capture the geometric relationship among different patches. Finally, the sequence of patch embeddings serves as input of the visual transformer encoder to pretrain.

With the rapid development of various ViT variants, there are also different ways to generate diverse patch features as in grid feature extraction: 1) \textit{Supervised Pre-training}: the image patch encoder is pre-trained with image classification data such as in ViT~\citep{dosovitskiy2020image} or object detection data such as in Swin Transformer~\citep{liu2021swin}; 2) \textit{Unsupervised Pre-training}: the image patch encoder is pre-trained with a large amount of unlabeled image-text pairs. For example, CLIP~\citep{radford2021learning} pretrains the image patch encoder of ViT with 400M aligned image-text pairs and \citep{changpinyo2021conceptual} provides a large dataset of 12M image-text pairs CC12M and conducts image-text pre-training to recognize long-tail visual concepts.

\subsubsection{Textual Features}
This research work utilizes the method in BERT~\citep{devlin2018bert} with the WordPiece tokenizer to tokenize the input text sentence into a sequence of sub-word tokens $\{w_1,\cdots,w_m\}$. Then each token $w_i$ is assigned three kinds of learnable embeddings: token, modal type and position embeddings. The three embeddings are summed and layer-normalized to represent input sentence as a sequence of embedding vectors $E_{emb}=\{e_{CLS},e_1,\cdots,e_m,e_{SEP}\}$, where $[CLS]$ and $[SEP]$ are the special tokens in BERT.

To provide better textual features, text stream parameters were initialized with three different pre-trained language models: BERT~\citep{devlin2018bert}, RoBERTa~\citep{liu2019roberta} and StructBERT~\citep{wang2019structbert}. RoBERTa trains on a larger corpus for more steps, and StructBERT incorporates more word ordering and sentence ordering information into pre-training language structures.

\subsection{Cross-Modal Interaction with Learning to Attend}
\input{vlp}
\input{l2a}

\subsection{Knowledge-guided Mixture of Experts}

Due to the complexity of the VQA task, there exist questions that are difficult to address by combining the diverse feature representation and V\&L pre-training. To address these questions and enable the model to evolve constantly, we further propose a knowledge-guided framework using the Mixture of Experts (MoE) model, as shown in Figure~\ref{framework}. Starting from a pre-trained V\&L model (the base Vision Understanding Expert in our case), a knowledge mining module is introduced to automatically discover the types of the questions that are not well-addressed by the Vision Understanding Expert, such as text-reading questions and clock-reading questions. These questions are then addressed by two extra expert modules newly introduced: Text Reading Expert and Clock Reading Expert, respectively. Finally, the three expert modules are combined together and routed to the right questions by the MoE model.

\subsubsection{Knowledge Mining}

On top of the comprehensive study of diverse feature representation and specific design of cross-modal interaction, we propose a continual learning framework to boost the power of the pre-trained V\&L model. It includes two stages: 1) identify new sub-tasks which require extra knowledge to learn; 2) learn expert models for these sub-tasks with the knowledge collected from domain experts or internet.

To identify new sub-tasks, we adopt a clustering-based method, which considers the low-confidence examples from a base model and discovers groups of these examples by their similarity to form new sub-tasks. Given a base model $M$ (i.e., Vision Understanding Expert in our case), we first collect examples which the model $M$ is difficult to give correct answers with high confidence. The model is unable to address these examples well with existing knowledge, which calls for specialized expert models with extra knowledge to handle them. Specifically, given an example $t$, the base model $M$ is designed to give a prediction with confidence score $s$. The output score on the predicted label of the Visual Understanding Expert is used to calculate this score $s$. The examples with low confidence scores ($s < 0.1$) thus indicate the cases that the base model finds difficult to handle. Then, the typical clustering algorithm K-Means \citep{macqueen1967some} is used to partition the set of these low-confidence examples into sub-task clusters. Under our V\&L circumstances, clustering is conducted on both the textual and visual content of examples. Therefore, the cross-modal representation of $[CLS]$ in the last layer of the Transformer are used as input to the clustering algorithm.

Clustering the low-confidence examples allows us to identify new sub-tasks.
In the VQA task, the clustering discovers the two types of visual questions (text-reading questions and clock-reading questions) that need OCR ability and clock-reading ability to deal with, respectively. Since the existing Vision Understanding Expert is incapable of resolving the two sub-tasks well, two extra expert modules: Text Reading Expert and Clock Reading Expert, are trained to deal with the low-confidence examples. Finally, the three expert modules work together as a complete VQA solution via the Mixture of Expert model.     

\subsubsection{Text Reading Expert}
\label{structlm}
\input{structlm}

\subsubsection{Clock Reading Expert}
\input{clock}

\subsubsection{Visual Understanding Expert}
Vison-and-Language Pre-training (VLP) models with different visual features can help capture diverse visual semantics from images, which facilitates deep vision-and-language understanding. For example, region features are good at capturing objects in an image, and thus very useful for answering questions on object counting. Grid features retain global or background information of an image, which can help to answer descriptive questions. Therefore, a \textit{diverse feature ensemble} method is used to construct our visual understanding expert, which ensembles multiple VLP models with different types of visual features: \textit{region feature}, \textit{grid feature} and \textit{patch feature}. For each kind of features, different VLP models are trained separately. A simple maximum voting strategy is then utilized to ensemble the different VLP models based on the prediction score. Our experiments demonstrate the advantage of the diverse feature ensemble over a single class of visual features.

\subsubsection{Putting It All Together by MoE}
\label{sec:moe}
\input{moe}

\section{Experiments}
\label{sec:exp}
\input{experiments}

\section{Related Work}
\label{sec:rel_work}
\input{rel_work}

\section{Discussion and Limitation}
\label{sec:discussion}
\input{conclusion}

\bibliographystyle{unsrtnat}
\bibliography{references}  






\end{document}

%% file: intro.tex
Recent years have witnessed human-level performance reached or surpassed by well trained computer programs in tasks ranging from games such as Go~\citep{alphago}
to classification of images in ImageNet~\citep{imagenet} to natural language understanding on the GLUE benchmark~\citep{glue}. Vision and language are two fundamental capabilities of human intelligence. We have seen dramatic progress in the area of representation learning across these two modalities. Inspired by the success of pre-training in both computer vision (CV)~\citep{razavian2014}
and natural language processing (NLP)~\citep{devlin2018bert}, a number of vision-and-language (V\&L) models have been proposed in the last couple of years to tackle challenges at the intersection of these two key areas of AI. Despite superhuman performance achieved respectively in some vision (e.g., ImageNet) and natural language tasks (e.g., GLUE), joint learning across these two modalities, which is essential to human cognition, has demonstrated limited human-level performance by prevalent V\&L approaches.

A compelling reason to study vision and language jointly is the promise of language as a universal and natural interface for visual reasoning problems - useful both in specifying a wide range of problems and in communicating AI responses. Visual reasoning has long been recognized most challenging for cross-modal learning owing to its requirement of higher-order cognition and commonsense reasoning intelligence. With the systematic design of our reasoning architecture, this research work unprecedentedly surpasses human performance in the popular Visual Question Answering (VQA) task~\citep{vqa}.


This paper summarizes how we achieve the human parity in VQA. Most of the presented techniques are not specific to VQA and can be transferable to tackling other visual reasoning tasks. Our work builds upon the significant progress made on CV and NLP over the past few decades.

\begin{figure} 
    \centering
    \includegraphics[width=0.5\textwidth]{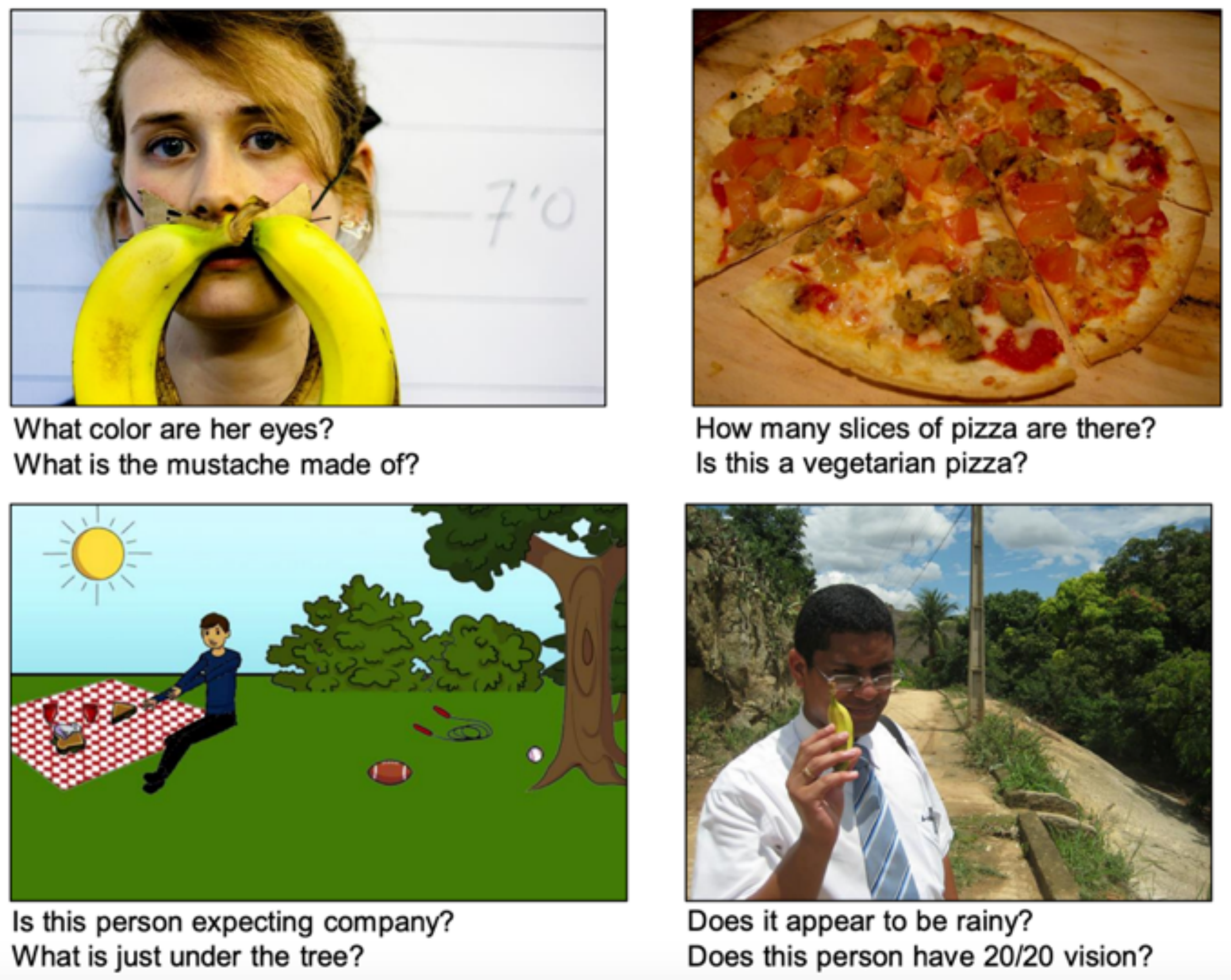}
    \caption{Examples of free-form, open-ended questions and images in the VQA dataset}
    \label{fig:vqa}
\end{figure}

A VQA system takes as input an image and a free-form, open-ended, natural-language question about the image, and produces a natural-language answer as the output. Example questions are shown in Figure~\ref{fig:vqa}. The open-ended questions require a potentially vast set of AI capabilities to answer, including question understanding, commonsense reasoning, activity recognition, object counting, and visually-grounded language understanding, etc.. Therefore, achieving human performance in VQA would be an important milestone in artificial intelligence. In pursuit of this goal, a new VQA architecture is designed by improving the individual capabilities of visual reasoning. Different from the prevalent VQA methods that rely on a single kind of features with standard Transformer, the new VQA architecture exploits more comprehensive visual and textual feature representation with pre-training, and more effective cross-modal interaction with \textit{learning to attend}.

The key to our success in VQA is tackling the diverse challenges with different capabilities. In particular, we introduce a novel knowledge mining framework with the Mixture-of-Experts (MoE) model for the complex VQA task. Most existing methods for VQA treat different types of visual questions in the same manner. Different from these methods, following the divide-and-conquer strategy, our new framework first decomposes the complex VQA task into different sub-tasks by a clustering-based method, which allows to identify the types of questions difficult to address. Each type of these questions are then resolved by a specialized expert module. All these expert modules are put together by the MoE paradigm. Beyond a patchwork of models like ensemble, the MoE paradigm learns which module to call upon by exploiting the expertise of each module, and thus intelligently delegates every question to a proper expert module. According to our quantitative analysis, the new knowledge mining with MoE plays an important role in boosting the performance of our VQA architecture up to the human level, significantly outperforming existing methods without explicit task decomposition by a large margin.


The rest of this paper is organized as follows. Section~\ref{sec:model} presents the design of our new VQA architecture with the knowledge mining framework. The empirical evaluation and quantitative analysis are given in Section~\ref{sec:exp}. Section~\ref{sec:rel_work} introduces the prior work related to VQA. Finally, the paper is concluded in Section~\ref{sec:discussion} by discussing our findings and limitations.

%% file: human_parity.tex
Despite immense progress on VQA in the research community over the past years, human parity has remained out of reach even though the gap is reduced significantly over the last a few years. This paper describes our efforts to achieve the unprecedented human-level performance on the VQA task by systematically improving the components of the VQA architecture, which is illustrated in Figure~\ref{fig:framework_overview}. This work addresses a number of limitations of existing VQA studies, and make the following major contributions:

\begin{figure}[b] 
    \centering
    \includegraphics[width=0.8\textwidth]{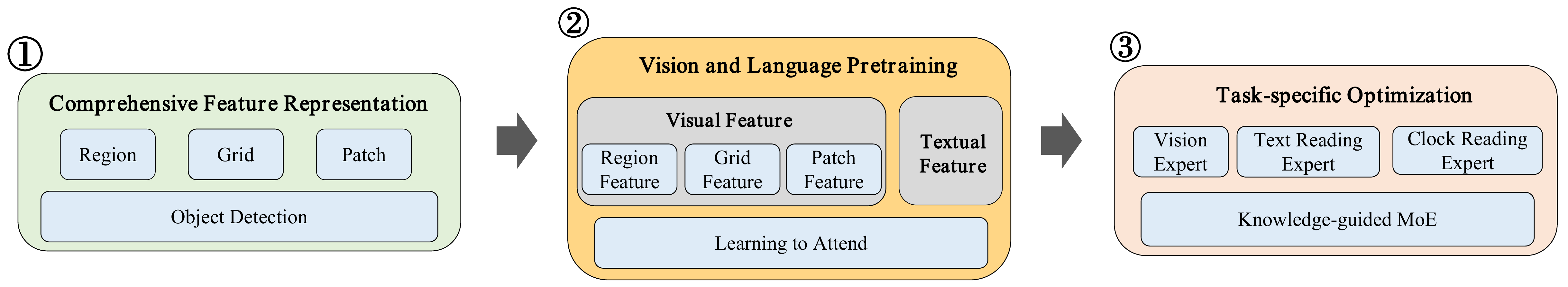}
    \caption{The overview of the new VQA architecture.}
    \label{fig:framework_overview}
\end{figure}

\begin{enumerate}
\item Most existing VQA approaches rely on a single class of features to represent visual signals. The homogeneous feature representation is insufficient to capture the diversity of visual signal needed to answer open-ended questions as illustrated in Figure~\ref{fig:vqa}. To address this limitation, our VQA architecture exploits a diverse set of visual representations: \textit{region features}, \textit{grid features} and \textit{patch features}, each of which is used to capture visual signals of a specific type. Specifically, \textit{region features} are good at locating the salient objects in the image, e.g., slices of pizza in the second case of Figure~\ref{fig:vqa}, which is more suitable in tasks like object counting. \textit{Grid features} are more skilled in the global or background information in the image, e.g., in the fourth case of Figure~\ref{fig:vqa}, where background information of weather is identified. With the heterogeneous feature set, the model is able to answer different types of questions by leveraging desired visual signals.

\item The semantic gap between visual and textual modalities has always been treated as one of the most important challenges in cross-modality research. There exist two families of V\&L models that bridge the cross-modal semantic gap: \textit{single-stream architecture}~\citep{su2019vl,chen2020uniter} and \textit{dual-stream architecture}~\citep{tan2019lxmert,yu2021ernie}. The single-stream architecture essentially treats the two input modalities equally, and thus does not make full use of the signal from each modality. On the other hand, the dual-stream architecture is insufficient to capture the fine-grained interaction between visual and textual hints. To address the limitations of these architectures, our VQA architecture fuses visual and textual modalities by learning how the features should attend to each other. To allow for fine-grained cross-modal interaction, our model is built upon the single-stream architecture. The original self-attention in Transformer is replaced with a weighted self-attention, where intra-modal attention and inter-modal attention are dynamically adjusted. This leads to effective alignment of the cross-modal semantics.

\item In the VQA task, visual questions are open-ended and require various kinds of knowledge and capabilities to answer. We, therefore, propose a new knowledge mining framework with MoE to address the diverse questions. Besides the general V\&L understanding and reasoning in VQA, the new framework is able to identify two types of questions (text-reading questions and clock-reading questions) that are difficult to address by the general-purpose VQA techniques, due to the lack of text reading and clock reading abilities. A specialized expert module is introduced for each of the two question types: 1) Text Reading Expert: answering questions by reasoning about text in images. 2) Clock Reading Expert: answering questions on the time shown by clocks.
An outstanding VQA architecture needs to appropriately mix the multiple experts to delegate each question to a proper expert module. Hence our VQA architecture employs the MoE paradigm to distill the expert knowledge required to answer each type of questions and mix the results of the experts to derive final answers. In the future, we will exploit techniques that automatically discover challenging case sets. New expert modules devoted to the discovered cases are then autonomously learned from specialized data. This learn-and-evolve process will lead to evolutionary intelligence that can rapidly adapt to any new task.

\end{enumerate}

%% file: vlp.tex
\subsubsection{Vision-and-Language Pre-training (VLP)}
The interaction of multiple modalities has always been treated as one of the most significant problems in cross-modality research. In VLP literature, there are two mainstream architectures to bridge the cross-modal feature gap: \textit{single-stream architecture} and \textit{dual-stream architecture}. The former such as VL-BERT~\citep{su2019vl} and UNITER~\citep{chen2020uniter} assume simple and clear underlying semantics behind the two modalities and thus simply concatenate image-region features and text features as input to a single Transformer~\citep{vaswani2017attention} network for early fusion in a straightforward manner. This paradigm learns the cross-modal semantic alignment from a bottom feature level by using the self-attention mechanism. Nevertheless, the design of single-stream structure treats both modality inputs equally, leaving the inherent different peculiarity of each modality not fully exploited. In contrast, the latter like LXMERT~\citep{tan2019lxmert} and ERNIE-ViL~\citep{yu2021ernie} first use separate Transformer encoders to learn high-level abstraction of image and sentence representation respectively, and then combine the two modalities together with a cross-modal Transformer. This kind of design explicitly distinguishes between different modality inputs and aligns the cross-modal representations at a higher semantic level, but is usually parameter-inefficient and may ignore fundamental feature-level association.

%% file: l2a.tex
\subsubsection{Learning to Attend}
\label{sec:l2a}
\citep{li2021semvlp} propose SemVLP to learn the joint representation of vision and language, which aligns cross-modal semantics at multiple levels. It builds upon a shared Transformer encoder with specific self-attention masks for cross-modal interaction. However, the interaction between the two modalities are controlled by fixed self-attention mask with only two modes: interactive or non-interactive. Therefore, our VQA architecture uses two learnable self-attention weights for each layer to dynamically control the inter-modal and intra-modal interaction.

In single-stream models, the input to a Transformer layer is the concatenation of both modalities, $X = [X_{L} | X_{V}]$. As a result, in each single-stream attention head, the query representation is given by:

\begin{align}
Q = XW^{Q}=\left(                 
  \begin{array}{c}   
    X_{L}  \\  
    X_{V} \\  
  \end{array}
\right)*W^{Q}= \left(                 
  \begin{array}{c}   
    Q_{L}  \\  
    Q_{V} \\  
  \end{array}
\right)
\end{align}

where $\left(                 
  \begin{array}{c}   
    \cdot L  \\  
    \cdot V \\  
  \end{array}
\right)$ are the language and visual sub-matrices
of the input and the resulting output. As shown in Figure \ref{framework}, the score matrix $S$ can be defined in terms of four sub-matrices:

\begin{align}
S = QK^{\mathrm{ T }} = \left(                 
  \begin{array}{c}   
    Q_{L} \\  
    Q_{V} \\  
  \end{array}
\right) \left(\begin{array}{cc} K_{L}^{\mathrm{ T }} & K_{V}^{\mathrm{ T }} \end{array}\right) = \left(
  \begin{array}{cc}   
    S_{LL} &  S_{LV} \\  
     S_{VL} & S_{VV} \\  
  \end{array}
\right) 
\end{align} 

Then, two learnable self-attention weights $\varepsilon_1$ and $\varepsilon_2$ are introduced for intra-modal attention score sub-matrices (diagonal of $S$) and inter-modal attention score sub-matrices (anti-diagonal of $S$), respectively. In each Transformer layer, the learnable weights are multiplied by the attention score matrix to obtain the new attention score matrix:

\begin{align}
S_{\gamma} = \left(                 
  \begin{array}{cc}   
    \varepsilon_1 S_{LL} & \varepsilon_2 S_{LV} \\  
    \varepsilon_2 S_{VL} & \varepsilon_1 S_{VV} \\  
  \end{array}
\right) 
\end{align}

The following two methods are investigated to learn the self-attention weights $\varepsilon_1$ and $\varepsilon_2$:
\begin{enumerate}

    \item The weights are derived from a single-layer feed-forward network with the sigmoid activation function. $V_{CLS}$ (the representation of $[CLS]$) is used as the input feature to reflect how well an image matches with text. It gives a useful signal to measure intra-modal and inter-modal interaction.
    \begin{align}
    (\varepsilon^1, \varepsilon^2) = FFN(V_{CLS})
    \end{align}
    
    \item The self-attention weights are learned directly as the two parameters with specified initial values:
    \begin{align}
    (\varepsilon^1, \varepsilon^2) = nn.Parameter(init\_value_1, init\_value_2)
    \end{align}

\end{enumerate}

\subsubsection{VLP with Learning to Attend}
Section~\ref{visual_fea} discusses the use of three classes of visual features (Region, Grid and Patch). Each class of the visual features and text feature are fused by the novel \textit{learning to attend} mechanism for cross-modal interaction, respectively.

\paragraph{Input Embeddings}
The input to Transformer is the image feature and its associated sentence (e.g. caption text). Each image is represented as a sequence of image features $\{o_{1},...,o_{n}\}$, and each sentence is represented as a sequence of words $\{w_{1},...,w_{m}\}$. The image and text embedding features are concatenated as input to the Transformer with learning to attend. The image representation for each kind of feature is described in Section~\ref{visual_fea}. To capture both local and global semantics of image and obtain diverse visual feature representation, different kinds of image features are fused by concatenating them together. It is then combined with the text embedding features as input to the Transformer with \emph{learning to attend}. This method is referred to as Fusion-VLP.


\paragraph{Pre-training Tasks}
The pre-training tasks of the three types (language, vision and cross-modality) are introduced in the pre-training stage, following LXMERT~\citep{tan2019lxmert}.
\begin{enumerate}
    \item \textbf{Masked LM Prediction.} The task setup is basically the same as in BERT~\citep{devlin2018bert}. The masked words are predicted by exploiting visual modality which helps to resolve ambiguity.
    
    \item \textbf{Masked Object Prediction.} Similarly, the vision side is pre-trained by randomly masking objects. In particular, 15\% of image objects are randomly masked, and the model is then asked to predict properties of these masked objects with the output object representations $O^L$.
    
    \item \textbf{Image-Text Matching (ITM).} This task randomly samples 50\% of mismatched image-text pairs and 50\% matched pairs, and trains a classifier to predict whether an image and a sentence match with each other on the representation.
    
    \item \textbf{Image Question Answering (QA).} The image question answering task is cast as a classification problem where the model is pre-trained with image QA data as in LXMERT~\citep{tan2019lxmert}. A classifier is then built on top of the representation $\textbf{h}^L_{CLS}$ in the model.
\end{enumerate}
For region-based features, the Region-VLP model is pre-trained with all the four pre-training tasks as in LXMERT~\citep{tan2019lxmert}, and the four losses are added up with equal weights. For grid feature, the Grid-VLP model is pre-trained with the pre-training tasks except \textit{masked object prediction}, since the grid feature does not capture explicit semantics. Besides, to accelerate the pre-training process, a random sampling strategy is adopted to dynamically sample 100 image grids for each image following PixelBERT~\citep{huang2020pixel}. The \textit{masked object prediction} task is also removed for Patch-VLP and Fusion-VLP.

During fine-tuning, the complete region/grid/patch features are used to retain all the extracted visual information. The hidden state $\textbf{h}^L_{CLS}$ of the last layer is used for cross-modality calculation.

%% file: structlm.tex
Text-reading VQA is an important VQA sub-task, which focuses on the questions that require to read the textual content shown in an input image. The existing models on the VQA dataset perform classification with frequent answers as labels. This classification modeling does not work well on text-reading samples where the answers are often not frequent enough to be included in the label set. Therefore, a specially designed deep LM that aims to capture structure information from texts, the StructuralLM model~\citep{structurallm} is utilized on these text-reading samples to extract answers from the text in images recognized by OCR. StructuralLM introduces cell-level 2D-positional embeddings and a new pre-training objective that classifies cells' positions. The pre-trained StructuralLM model is adopted to the text-reading VQA samples in the following way.

In order to adapt our StructureLM to image texts, we fine-tuned a pre-trained StructuralLM by text-reading samples. In particular, an OCR tool is first used to recognize text and serialize the cells (bounding boxes) from top-left to bottom-right in images. Each image is represented as a sequence of cells $\{c_{1},...,c_{n}\}$, each of which contains a sequence of words $c_i=\{w_{i}^{1},...,w_{i}^{m}\}$. A separator $[SEP]$ is added between every two bounding boxes to separate them, which gives an input sequence $\{q_{1},...,q_{e},[SEP],c_{1},[SEP],c_{2}..,[SEP],c_{n}\}$. The StructuralLM is pre-trained subsequently in the same way as it is pre-trained on document images. A token-level span prediction classifier is then built upon the token representation to perform an extractive QA task, as often did for machine reading comprehension~\citep{rajpurkar2016squad,chen2017reading}. Finally, the added separator is removed from the predicted answer span.

%% file: clock.tex
With the powerful VQA features aforementioned, many questions can get satisfactory answers. However, it still suffers from reading precise time from clocks, as clock reading requires specific prior knowledge. Hence, a clock reading expert is introduced to address such kind of problems.
The clock reading expert consists of a clock detector and a clock reader. The clock detector is used to detect clocks in images, which is essentially an object detection task. The Cascade-RCNN~\citep{cai2018cascade} is used as the backbone network for the clock detector. A binary classification loss and a bounding box regression loss are applied for training as the standard detection framework does~\citep{cai2018cascade}.
The detected bounding boxes from the clock detector are fed into the clock reader, which reads the precise time in the clocks. The clock reading is modeled as both a classification task as well as a regression task.

%


In the clock reader, Resnet50-IBN~\citep{pan2018two} is adopted as our backbone, and two specific branches are introduced for hour and minute prediction respectively. Furthermore, as the hour and minute hands in the clock are the keys to predict the time, attention modules were introduced to force the focus of the model on the hands.
A SE-layer~\citep{hu2018squeeze} is employed after the backbone for channel-wise attention, and a spatial attention module which consists of convolution layers and ReLU activation is employed in the beginning of hour and minute branch respectively for spatial-wise attention. Such a corporation of channel and spatial wise attention is able to adapt to the individual bias of hour and minute prediction. The feature outputs from two branches are listed as following:
\begin{equation}
f_{m} = E_m(Attn_{sp}(F)*F), \qquad f_{h} = E_h(Attn_{sp}(F)*F)
\end{equation}
where $I$ is the image, and $E$, $E_h$, $E_m$ are the backbone, hour branch and minute branch respectively. $F$ is the feature map from the backbone after SE-layer, $F = Attn_{se}(E(I))$. 

As the clock reader is formulated as both a classification task and a regression task, it introduces loss from two perspectives. From the classification aspect, a 12-category classification loss is used for both hour and minute \footnote{the minute task is divided into 12 bins by 5 moves per bin.} prediction. The cross-entropy loss is adopted as follows: 
\begin{equation}
L_{cls} = - \sum^N_{i}  g_{i} \times \log p_{i} 
\end{equation}
where $N$ is the number of categories and set to 12.



As 2:00 is closer to 3:00 than to 9:00, it is also important to solve the problem from a regression perspective. The regression loss is listed as below:
\begin{equation}
L_{reg} = \sum^B_{i} \left ( cos\left ( \frac{2\pi}{C} \times \left ( p_{i}-g_{i} \right ) -\pi \right ) + 1 \right )
\end{equation}
where $B$ is the batch size. The cosine formulation is used for the periodicity constraint of the clock prediction. $C$ is the periodicity of hour or minute, which is set to 60 moves. $p_{i}$ and $g_{i}$ is the prediction and ground truth.

Because one full turn of the minute hand corresponds to 5 moves of the hour hand, a self-supervised loss is introduced. The self-supervision of hour and minute is regarded as a regularization loss to improve the generalization of clock reader.
\begin{equation}
  L_{self} = \sum^B_{i}Smooth_{L1}\left ( C \times \left ( p_{h}-\left [ p_{h} \right ] \right )-p_{m} \right )
\end{equation}


Finally, the total loss is: 
\begin{equation}
    L = L_{cls} + L_{self} + \lambda L_{reg}
\end{equation}
where $\lambda$ is used to weight the self-supervised loss, and set to 0.01.

%% file: moe.tex
The methodology of Mixture of Experts (MoE) ~\citep{jacobs1991adaptive,shazeer2017outrageously} essentially decomposes a task into sub-tasks, and develops an expert on each sub-task. A gating network is then used to coordinate multiple experts for task completion. We follow the recent work of Switch Transformers~\citep{fedus2021switch}, and adopt the simple routing strategy that the the gating network routes to only a \emph{single} expert. 
We use the simplest form to preserves model quality and reduce routing computation.
In our framework, the VQA task can be decomposed into three sub-tasks according to the analysis of the visual questions via proposed knowledge mining. A multi-layer perception network is trained as the gating network, which performs three-class classification to determine which expert to choose for each instance.  

Given an expert $M_t$ for sub-task $t$, the expert will give an answer and we compute a reward score $s_t$ between the predicted answer and the human annotated labels using Equation \ref{eq:evaluation}. The reward score $s_t$ is used as supervision for training, where the network is trained to route each instance to its best-match expert.  At training time, we propose to maximize the Binary Cross Entropy (BCE) loss $L$ as follows:
\begin{equation}
    L_{MoE} = \sum_{t}s_t\log{\hat{s_t}} + (1-s_t)\log{(1-\hat{s_t})}
\end{equation}
where $s_t$ denotes the ground-truth reward score of sub-task $t$, $\hat{s_t}$ stands for the prediction score of the MoE network. At test time, we choose the single routed expert with the maximum prediction score $\hat{t} = arg\max \hat{s_t}$. The prediction score $\hat{s}$ is calculated using a Multi-Layer Perception network (MLP) as follows:
\begin{equation}
    \hat{s} = W_3(\tanh(W_2\tanh(W_1 x + b_1) + b_2)) + b_3
\end{equation}
where $x$ is the input feature and $W_i, b_i$ are the learnable parameters.

The following features are derived for training the gating network:
\begin{itemize}
    \item \textit{Each expert's confidence}: for Visual Understanding Expert, the maximum prediction score is used for confidence score. For Text Reading Expert and Clock Reading Expert, the output score is used for confidence score, and if an image does not have text or any clock, the score is set to $-1$.
    \item \textit{Question type}: A three-class classifier is trained to predict whether a question is asked about text reading, clock reading or visual understanding. To train the classifier, OCR-labeled data is collected from TextVQA~\citep{textvqa} \& STVQA~\citep{stvqa}, and clock-labeled data from the VQA dataset by retrieving the keywords \emph{clock} and \emph{what time}. Other cases from VQA data are sampled as negative samples by the ratio of 1:2. The prediction scores of these three classes are used as the input features.
\end{itemize}
Even though the current process is manually designed, in the future, we will be exploiting techniques that allow us to automatically discover subset of challenging cases, and incrementally add more experts to address the discovered cases. 




%% file: experiments.tex
\subsection{Data}

\paragraph{Pre-training Data}
The same in-domain data is used as in LXMERT~\citep{tan2019lxmert} for pre-training. It consists of the image caption data from MS COCO~\citep{lin2014microsoft}, Visual Genome~\citep{krishna2017visual}, image question answering data from VQA 2.0~\citep{antol2015vqa}, GQA balanced version~\citep{hudson2019gqa} and VG-QA~\citep{zhu2016visual7w}. The total amount of the dataset is 9.18M image-and-sentence pairs on 180K distinct images. Also, additional out-of-domain data from Conceptual Captions~\citep{sharma2018conceptual} for model pre-training, which consists of about 3M image-text pairs on 3M images.

For Text Reading Expert, the StructuralLM~\citep{structurallm} is used as the base model, which is pre-trained on the IIT-CDIP Test Collection 1.0~\citep{Lewis}. It is a large-scale scanned document image dataset containing more than 6 million documents, with more than 11 million scanned document images.

\paragraph{Fine-tuning Data}
Visual Question Answering (VQA) is a dataset containing open-ended questions about images~\citep{antol2015vqa}. These questions require understanding of vision, language and commonsense knowledge to answer. It contains a large number of labeled question-image-answer triplets with 10 human annotators for each question. The detailed statistics for VQA training/validation/test data splits is shown in Table~\ref{tab:vqa_stat}. 

\begin{table}[t]
\caption{VQA Data Statistics.}
\centering
\begin{tabular}{c|c|cccc|c}
\toprule
 & Images & Questions & Yes/No & Number & Other & Answers  \\
\midrule
Training                                                    & 80K & 443K & 169K & 58K & 219K & 4.4M  \\
Validation                                                 & 40K & 214K & 81K & 28K & 106K & 2.1M  \\
Test & 80K    & 447K & - & - & - & - \\
\bottomrule
\end{tabular}
\label{tab:vqa_stat}
\end{table}

For training Text Reading Expert, three text-reading VQA datasets is used including a subset of VQA data~\citep{antol2015vqa}, TextVQA~\citep{textvqa} and ST-VQA~\citep{stvqa}. A classification model is trained to extract text-reading samples from VQA data. The questions of TextVQA and ST-VQA are treated as positive samples, and the questions on images without text in VQA are treated as negative samples. The detailed statistics for the three text-reading VQA datasets is shown in Table \ref{tab:textvqa_stat}.

\begin{table}[ht]
\caption{\label{table:compare} Text-reading VQA Data Statistics.}
\centering
\begin{tabular}{l|c|c}
\toprule 
\textbf{Dataset} & \textbf{Images} & \textbf{Questions} \\
\midrule
VQA-Subset & 20k & 21k  \\
TextVQA & 25k & 39k  \\
ST-VQA & 19k & 26k  \\
\bottomrule
\end{tabular}
\label{tab:textvqa_stat}
\end{table}

For training Clock Reading Expert, the images are collected from two sources. One is from \textit{open-access datasets}. Specifically, a total of 4863 images are collected from COCO2017~\citep{lin2014microsoft}\footnote{The COCO images used in the VQA test set are left unlabeled and excluded from our training data.} and 2691 images from ImageNet~\citep{deng2009imagenet} for clock labeling, both of which are widely used open-access datasets. Annotators are required to give the bounding boxes and the precision time of clocks in images. After labeling, 4236 and 3271 valid clock bounding boxes are obtained from COCO2017~\citep{lin2014microsoft} and ImageNet~\citep{deng2009imagenet} respectively. 785 clock bounding boxes are randomly sampled from COCO2017 images for validation.
The other source is \textit{Internet images}. To further increase the generalization and capacity of our clock reader, 2878 images from internet with various clocks are collected. After careful annotation, 2314 valid clocks are obtained. Note that this data is only used for the training of the clock reader.

\paragraph{Evaluation Metric}
Following \citep{antol2015vqa}, an evaluation metric robust to inter-human variability is used in phrasing the answers: 
\begin{equation}\label{eq:evaluation}
\text{Acc({\it ans})} = \text{min} \Big\lbrace \frac{\text{\# human that said {\it ans}}}{3}, 1 \Big\rbrace
\end{equation}
In order to be consistent with ``human accuracies'', machine accuracies are averaged over all 10-choose-9 sets of human annotators.

\subsection{Experimental Setup}
\paragraph{VLP} The maximum sequence length for the sentence is set as 20. For the VLP models, the pre-trained Transformer encoder with 12 layers is used as our base architecture, and the one with 24 layers as the large architecture. The basic settings of the Transformer are the same as BERT~\citep{devlin2018bert}, and the Transformer encoder is initialized with StructBERT~\citep{wang2019structbert} for its good performance. For the method of \emph{learning to attend}, the two learnable parameters are initialized with $init\_value_1=1.0$ and $init\_value_2=L_s/L$, where $L$ is the number of Transformer layers and $L_s$ is the corresponding layer number. The base model is pre-trained with a total batch size of 512 for 30 epochs on 8 A100 GPUs and the AdamW optimizer with the initial learning rate of 1e-4. The 24-layer large architecture is pre-trained with the total batch size of 512 on 8 A100 GPUs. To deal with over-fitting, two-stage pre-training strategy is employed as in LXMERT~\citep{tan2019lxmert}. Specifically, the model is first pre-trained without the question answering task with the initial learning rate of 5e-5 for 20 epochs, and then pre-trained with all the tasks together with the initial learning rate of 2e-5 for another 10 epochs. The detailed settings for the three VLP methods are listed as below:
\begin{enumerate}
    \item \textbf{Region-VLP}: The detection model is used in VinVL~\citep{zhang2021vinvl} to detect objects and extract region features. It is a large-scale object-attribute detection model based on the ResNeXt-152 C4 architecture. 100 objects is retained for each image to maximize the pre-training compute utilization by avoiding padding.
    
    \item \textbf{Grid-VLP}: It follows the basic settings in Grid-VLP~\citep{yan2021gridvlp}
    . ResNeXt is chosen to be the visual encoder with different sizes~\citep{xie2017aggregated} as in~\citep{jiang2020defense,huang2020pixel}. The shorter side of every input image is resized to 600, and the longer side is limit to at most 1000. A fixed number of 100 grids are randomly selected each time during pre-training~\footnote{\small{We also tested with 64 and 128 selected grids. It did not lead to significantly different results.}}. 
    
    \item \textbf{Patch-VLP}: It uses the visual Transformer encoder of the Swin detector~\citep{liu2021swin} and CLIP~\citep{radford2021learning}. The ViT-B/32 pre-trained model is chosen, which has 12 Transformers layers with input patches of size $32 \times 32$. Every input images is resized to $224 \times 224$ as CLIP does, resulting in $7\times7=49$ patches.
    
    \item \textbf{Fusion-VLP}: It fuses the three classes of image features (Region, Grid and Patch) by concatenating them together as the visual input to the Transformer. A two-stage strategy is employed to pre-train Fusion-VLP, which first trains the region-grid model initialized with Region-VLP, and then continues to train the region-grid-patch model.
\end{enumerate}

\paragraph{Fine-tuning on VQA}
Following~\citep{anderson2018bottom}, our architecture treats VQA as a multi-class classification task by picking an answer from a shared set of 3,129 answers. The hidden state of $h^L_{CLS}$ is used to map the representation into 3,129 possible answers with an additional MLP layer. The model is trained with a binary cross-entropy loss on the soft target scores. The pre-trained models are fine-tuned based on the three classes of features on the VQA training data for 3 epochs with the batch size of 32, and the BERT Adam optimizer is employed with the initial learning rate of 1e-4 for base models and 2e-5 for large models. At inference, a \emph{softmax} function is used for prediction.

\paragraph{Text Reading Expert} 
The text reading expert follows the basic settings in StructuralLM ~\citep{structurallm} and uses the pre-trained StructuralLM-large as the backbone model. In particular, StructuralLM is pre-trained with a batch size of 16 for 50K steps. The question tokens and the OCR tokens of an image are concatenated as an input sequence, of which the maximum length is set as 128. For fine-tuning, the three kinds of text-reading VQA datasets are merged and split with 10-fold cross-validation. The StructuralLM is fine-tuned with the total batch size of 16 for 4 epochs, and the AdamW optimizer is employed with the initial learning rate of 3e-5. Accuracy and ANLS (Average Normalized Levenshtein Similarity) are used as the metrics to evaluate the text reading expert.

\paragraph{Clock Reading Expert}
The clock detector of the clock reading expert is trained following the basic settings of Cascade-RCNN~\citep{cai2018cascade}. The clock reader is trained with the batch size of 96 in 2 GPUs, and the initial learning rate is set as 0.02. It is trained for 150 epochs with the learning rate multiplied by 0.1 at 90-th and 120-th epochs. The data augmentation pipeline consists of $256 \times 256$ random resized cropping, random color jittering, random gray-scale conversion, Gaussian blurring and random rotation within $\pm 45^{\circ}$.

\paragraph{Visual Understanding Expert} 
The visual understanding expert ensembles 46 models in total, including 14 Region-VLP models, 21 Grid-VLP models, 4 Patch-VLP models and 7 Fusion-VLP models. Simple maximum voting is adopted to ensemble all the models based on their prediction scores.

\paragraph{Mixture of Experts} 
The MoE adopts Multi-layer Perceptron (MLP) as the gating network to determine experts for given questions. The MLP has two hidden layers of 100 neurons and 50 neurons, respectively. It uses \emph{tanh} as the activation function, and the Adam optimizer with the initial learning rate of 1e-3. The network is trained for 5 epochs with the batch size of 256.


\begin{table}[t]
\caption{VQA Challenge Leaderboard.}\label{tab:main_results}
\centering
\begin{tabular}{l|c|ccc}
\toprule
\multicolumn{5}{c}{VQA Challenge Leaderboard (Test-std)}                                                             \\
\midrule
Models                                                    &  Overall & Yes/No & Number & Other \\
\midrule
Human    &  80.83 & \bf 95.48 & \bf 81.29 & 67.97\\
\midrule
LXMERT (\cite{tan2019lxmert})                   & 74.34    & 89.45  & 56.69  & 65.22 \\
MCAN (\cite{yu2019mcan})                      & 75.23    & 90.36  & 59.17  & 65.75 \\
VILLA (\cite{gan2020large})                     & 75.85    & 91.30  & 59.23  & 66.20  \\
BGN (\cite{guo2019bilinear})                      & 75.92    & 90.89  & 61.13  & 66.28 \\
InterBERT (\cite{lin2020interbert})          & 76.10    & 91.67  & 59.24  & 66.40  \\
GridFeat+MoVie (\cite{jiang2020defense})           & 76.29    & 90.81  & 61.53  & 67.04 \\
VinVL (\cite{zhang2021vinvl})                     & 77.45    & 92.38  & 62.55  & 67.87 \\
ROSITA (\cite{cui2021rosita})                   & 78.34    & 92.66  & 63.24  & 69.33 \\
UNIMO (\cite{li2020unimo})  & 78.40  & 93.10  & 63.06  & 69.12 \\
VQA Challenge 2021 winner & 79.34    & 93.28  & 65.36  & 70.40  \\
PASH-SFE                  & 79.47    & 92.45  & 76.57  & 68.82 \\
SimVLM (\cite{wang2021simvlm})                   & 80.34    & 93.29  & 66.54  & 72.23 \\ 
\midrule
\modelname                 & \textbf{81.26}    & 93.55  & 72.01  & \bf 72.67 \\
\bottomrule
\end{tabular}
\end{table}

\begin{table}[b]
\setlength{\tabcolsep}{0.6mm}
\caption{Performance comparison with other single models.} \label{tab:single_results}
\centering
\begin{tabular}{lc|ccc|ccc}
\toprule
\multicolumn{8}{c}{Performance of Single Models}                                                                                           \\
\midrule
\multirow{2}{*}{Models}                             & \multirow{2}{*}{Feature Type} & \multicolumn{3}{c|}{BASE}     & \multicolumn{3}{c}{LARGE}    \\
                                                    &                        & Params & Test-dev & Test-std & Params & Test-dev & Test-std \\
\midrule
VLBERT (\cite{su2019vl})           & Region                 & 110M   & 71.16    & -        & 345M   & 71.79    & 72.22    \\
UNITER (\cite{chen2020uniter})     & Region                 & 110M   & 72.70    & 72.91    & 345M   & 73.82    & 74.02    \\
OSCAR (\cite{li2020oscar})         & Region                 & 110M   & 73.16    & 73.44    & 345M   & 73.61    & 73.82    \\
UNIMO (\cite{li2020unimo})         & Region                 & 110M   & 73.79    & 74.02    & 345M   & 75.06    & 75.27    \\
VinVL (\cite{zhang2021vinvl} )     & Region                 & 110M   & 75.95    & 76.12    & 345M   & 76.52    & 76.60    \\
ViLBERT (\cite{lu2019vilbert})     & Region                 & 221M   & 70.55    & 70.92    & -      & -        & -        \\
12-in-1 (\cite{Lu_2020_CVPR})    & Region                 & 221M   & 73.15    & -        & -      & -        & -        \\
LXMERT (\cite{tan2019lxmert}  )    & Region                 & 183M   & 72.42    & 72.54    & -      & -        &          \\
ERNIE-ViL (\cite{yu2021ernie} )    & Region                 & 250M   & 73.18    & 73.36    & 510M   & 74.95    & 75.10    \\
PixelBERT (\cite{huang2020pixel}) & Grid                   & 170M   & 74.45    & 74.55    & -      & -        & -        \\
ViLT (\cite{kim2021vilt}) & Patch                   & 110M   & 71.26   & -    & -      & -        & -        \\
\midrule
Region-VLP                & Region                                            & 110M   &  76.25 &  -  & 345M &  77.17 &  - \\
Grid-VLP                  & Grid                                              & 110M   &  76.50 &  -  & 345M &  77.13 &  - \\
Patch-VLP                 & Patch                                             & 110M   &  71.61 &  -  & 345M &  - &  - \\
Fusion-VLP                                           & Region+Grid+Patch      & 110M   & \textbf{76.80}        & \textbf{76.78}        & 345M   & \textbf{77.59}   & \textbf{77.61}  \\ 
\bottomrule
\end{tabular}
\end{table}

\subsection{Main Results} \label{sec:mr}


Table~\ref{tab:main_results} presents our main results compared with all the previous public and unpublic best results on the VQA Challenge Leaderboard. From the results, it can be observed that: 1) Our VQA architecture \modelname represents the first to achieve human parity on VQA Challenge Leaderboard outperforming all the previous state-of-the-art methods by a large margin, which demonstrates the effectiveness of our framework. 2) With regard to a breakdown of performance on different question types, \modelname performs much better on the ``Other'' type than human do, and gives comparable results on ``Yes/No'' questions. \modelname performs worse than human do on type ``Number'' for the two reasons: a) in the ``Number'' type, there are many questions about reading OCR text, which are easier for human to answer; and b) there are many object counting questions that are more difficult for \modelname to answer.

Table \ref{tab:single_results} presents the detailed results of our single VLP models compared with other state-of-the-art methods. From the results, it is observed that: 1) the proposed VLP model outperforms the others on every kind of visual feature (region / grid / patch), respectively. It demonstrates the effectiveness of the proposed cross-modal interaction with learning to attend mechanism. 2) The methods with self-attention on patch feature perform worse than the region-based and grid-based methods do. There are two weaknesses of patch-based methods: a) the visual semantic information is not well-captured in existing patch-based VLP methods. How to inject visual semantics into patch representation remains largely unexplored; b) the image-text pre-training data is not enough for large-scale patch-based pre-training; 3) Fusion-VLP gives the best performance by fusing all the three classes of visual features as input, which validates the effectiveness of comprehensive feature representation.

\subsection{Model Analysis by Modules} 
\label{sec:model_analysis}

\paragraph{Visual Feature Importance}
Here presents the ablation study to assess the importance of different visual features for VLP on the VQA test-dev set. The results shown in Table \ref{table:vf} indicate that: 1) The VLP methods based on region and grid features achieve much better performance than the ones based on patch feature do, as stated in Section~\ref{sec:mr}. When examining by individual question types, Region-VLP performs better on the ``Number'' type, while Grid-VLP does better on the ``Yes/No'' and ``Other'' types. The difference can be attributed to the fact that region feature captures more local information of an image at the object level, and thus is more effective in address the visual counting problem by identifying local objects in an image. On the other hand, grid feature captures globally visual context in an image, which helps to answer the ``Yes/No'' and ``Other'' questions; 2) by combining the three classes of features in the way of early fusion, Fusion-VLP performs the best among all the single models. It shows that the different kinds of features can complement well with each other.

\begin{table}[t]
\centering
\caption{Ablation study of visual features on VQA Test-dev.} \label{table:vf}
\begin{tabular}{lcccc}
\toprule
           & Overall & Yes/No & Number & Other \\
\midrule
Fusion-VLP &  \textbf{77.59}    &  91.91    &   \textbf{64.29}    &    \textbf{68.33}   \\
Region-VLP & 77.17 & 91.62  & 63.69 &  67.84     \\
Grid-VLP   &  77.13  &  \textbf{92.20}    &   59.99     &   68.15    \\
Patch-VLP  &    71.61     &  88.17      &   49.44     &   62.54    \\
\bottomrule
\end{tabular}
\end{table}

\paragraph{Learning to Attend}
Here presents the ablation study to assess the importance of the \emph{learning to attend} mechanism on the VQA test-dev set. The 24-layer Region-VLP is used as the baseline model, which is pre-trained and fine-tuned based on the original Transformer. The ablation applies the same pre-training and fine-tuning settings, and only modifies the self-attention block with the two ways of \emph{learning to attend} stated in Section~\ref{sec:l2a}. From Table \ref{table:l2a}, it can be seen that the model with either way of \emph{learning to attend} outperforms the best Region-VLP baseline. Among the two different ways, the one with two learnable parameters performs slightly better than the other one. The reason may lie in: 1) learning two unrestricted parameters for each layer allowing for more parameter freedom, so as to better align cross-modal semantics, 2) the discrepancy between pre-training and fine-tuning, where the representation of [CLS] is learnt to model the semantic relation between a caption and an image during pre-training, while it is repurposed for question answering in fine-tuning.


\begin{table}[b]
\centering
\caption{Ablation study of \emph{learning to attend} on VQA Test-dev.}
\begin{tabular}{lcccc}
\toprule
                  & Overall & Yes/no & Number & Other \\
\midrule
Region-VLP (Baseline)       &   76.75 & 91.28 &  63.31 &  67.34  \\
\quad + Learning to Attend (FFN) &   77.09 &  91.58 &  63.54 & 67.74  \\
\quad + Learning to Attend (Param) &  \textbf{77.17} & \textbf{91.62}  & \textbf{63.69}  &  \textbf{67.84} \\

\bottomrule
\label{table:l2a}
\end{tabular}
\end{table}

\begin{table}[!h]
\centering
\caption{Ablation study of text reading expert on the VQA Test-dev.}
\begin{tabular}{lcccc|c}
\toprule
           & Overall & Yes/No & Number & Other & ANLS \\
\midrule
Visual Understanding Expert & 79.44  & 93.31 &  65.70 & 71.16 & - \\
 \: + Text-reading VQA data & 80.35 & 93.31 & 69.81 & 71.49 & 79.85 \\
 \: \: + add separator &  80.41 &   93.31  & 69.82   &   71.64 & 79.96 \\
 \: \: \: + continue pre-training  & \textbf{80.63} & 93.31 & \textbf{69.97} & \textbf{72.01} & \textbf{80.33} \\
\bottomrule
\label{tab:structlm}
\end{tabular}
\end{table}

\begin{figure} 
    \centering
    \includegraphics[width=\textwidth]{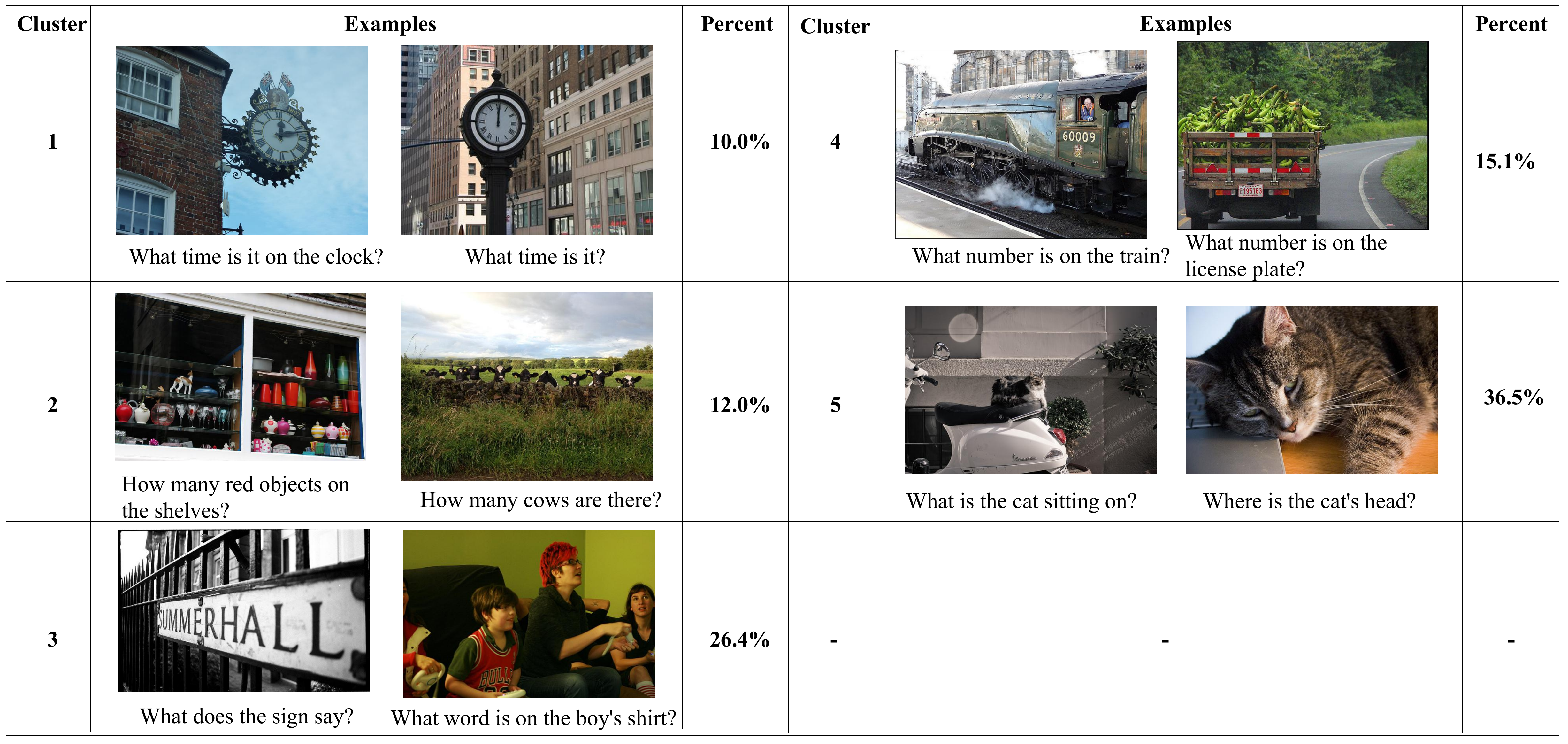}
    \caption{Illustration examples of clustering results. Percent is the percentage number of each cluster's examples.}
    \label{fig:kmeans_example}
\end{figure}

\begin{figure} 
    \centering
    \begin{subfigure}[]{}
         \centering
         \includegraphics[width=0.45\textwidth]{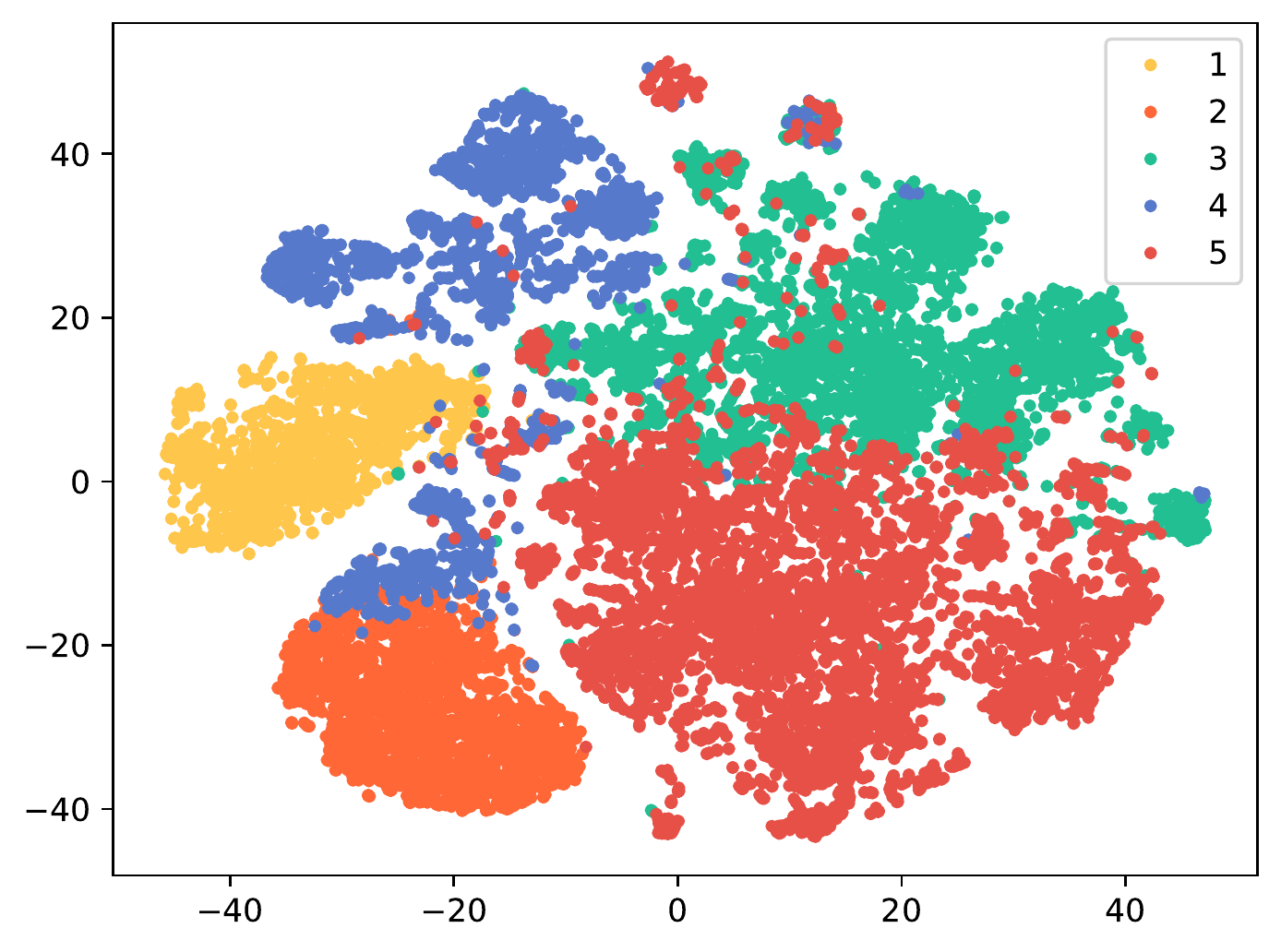}
     \end{subfigure}
     \hfill
     \begin{subfigure}[]{}
         \centering
         \includegraphics[width=0.45\textwidth]{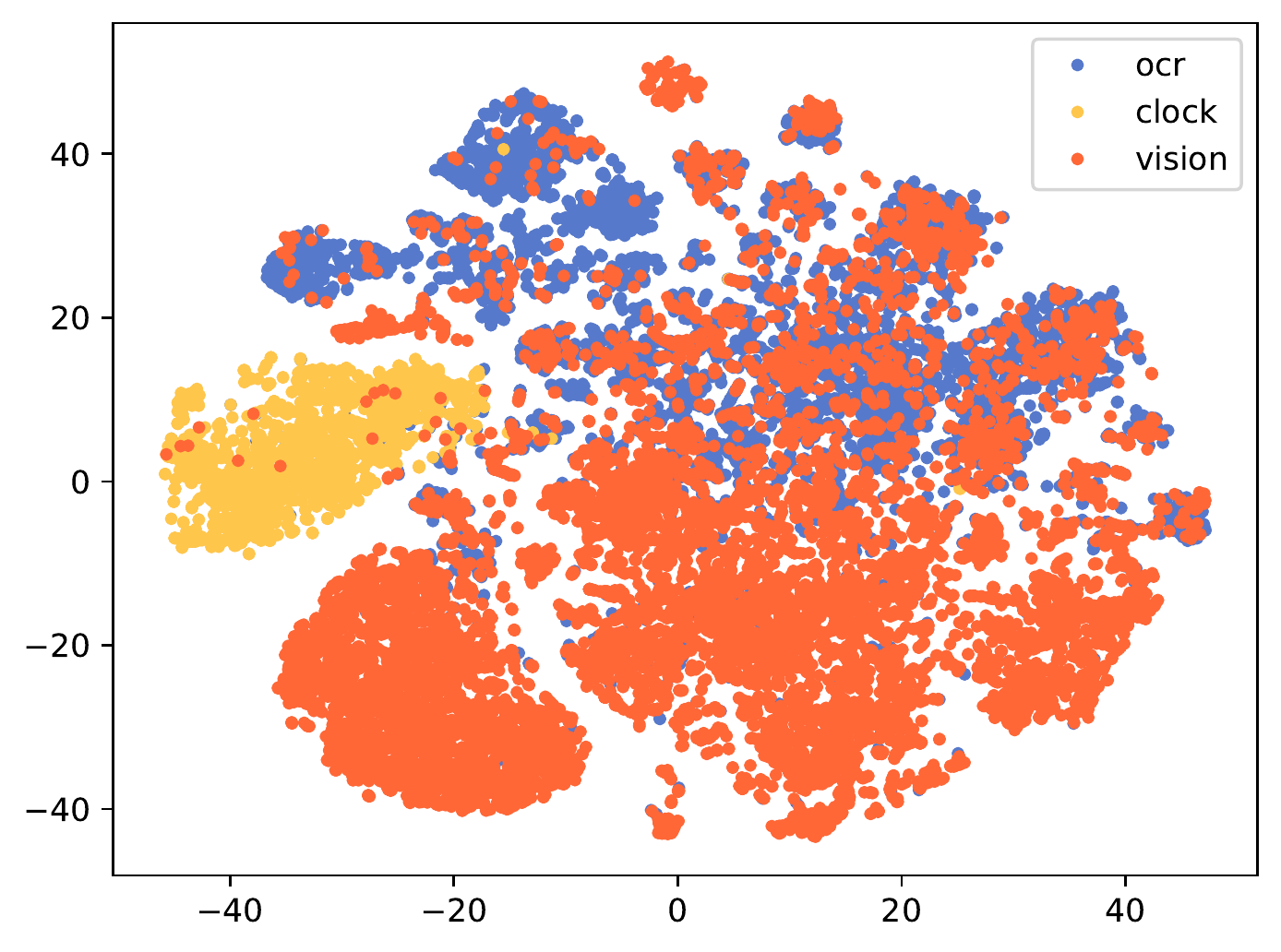}
     \end{subfigure}
    \caption{The t-SNE visualization of clustering results. Figure (a) shows the clustering results and the label 1/2/3/4/5 is cluster id. Figure (b) shows the classification results and the label ocr/clock/vision is classification label. The classifier is manually built and the accuracy of it is 95.0\%.}
    \label{fig:tsne}
\end{figure}

\paragraph{Knowledge Mining}
Figure~\ref{fig:kmeans_example} illustrates the clustering result. We choose the number of the clusters as 5, which gives the best performance on our quantitative test. For each cluster, we showcase two examples in the cluster and the percentage of the examples in this cluster. From the results, we can see that the proposed knowledge mining can actually mine certain meaningful topic clusters, where similar examples are clustered together. For example, Cluster~1 is about asking questions about time and clocks. Cluster~2 is about counting problems. Cluster~3 is about reading texts from the wall or the clothes. Cluster~4 is about reading number texts on the vehicles. We found that there are two new tasks: clock-reading task (Cluster~1) and text-reading task (Cluster~3,4), both of which require specific prior knowledge. We also use a three-class classifier in Section~\ref{sec:moe} to classify the filtered candidate examples, so as to measure the consistency between the clustering result and classification result.
Figure~\ref{fig:tsne} gives t-SNE visualization of the clustering result and classification result. From the result, we can see high consistency between the clustering and classification result on the measured topics. The knowledge mining method can properly separate part of the clock-reading examples and OCR-reading examples from the other examples, although for the OCR-related exampled there still exist limited examples mixed up in the common vision category.

To measure the consistency of the clustering result to the classification labels, we also provide detailed quantitative analysis on different clustering methods. We manually build a three-label classifier (OCR, clock and vision) with 95\% accuracy as in Section~\ref{sec:moe} and apply it to evaluate the consistence of each cluster. We project each cluster to the corresponding label heuristically. For example, in Figure~\ref{fig:kmeans_example}, Cluster~1 is assigned to clock label, Cluster~3 and Cluster~4 are to OCR label, and Cluster~5 is to vision label. We then compare the assigned label of each cluster to that of the classification label (95\% accuracy). We use accuracy, macro-precision, macro-recall and macro-f1 to measure how consistent the compared label in each cluster is. As list in Table~\ref{tab:cluster}, K-Means (K=5) achieves the best performance with 0.8448 accuracy and 0.8739 macro-F1, which shows that the clustering result is highly consistent with the assumed classification labels on OCR/clock/vision. 


\begin{table}[h]
\centering
\caption{Quantitative analysis on the clustering results of different clustering methods.}\label{tab:cluster}
\begin{tabular}{l|c|ccc}
\toprule
                  & Acc    & P      & R      & Macro-F1 \\ 
\midrule
DBSCAN (eps=0.5)  & 0.1544 & 0.4605 & 0.3861 & 0.1466   \\
K-Means (K=3)       & 0.4969 & 0.6487 & 0.662  & 0.6163   \\ 
K-Means (K=4)       & 0.7894 & 0.8239 & 0.8219 & 0.8195   \\ 
K-Means (K=5)       & \textbf{0.8448} & 0.8659 & 0.8898 & 0.8739   \\ 
K-Means (K=6)       & 0.8443 & \textbf{0.8668} & \textbf{0.8918} & \textbf{0.8740}   \\ 
\bottomrule
\end{tabular}
\end{table}

\paragraph{Text Reading Expert}
As stated in Section~\ref{structlm}, the pre-trained StructuralLM model is adapted on text-reading VQA samples in different ways. Table \ref{tab:structlm} shows the ablation results on the VQA test-dev set. The visual understanding expert~\footnote{\small An early version of visual understanding expert for VQA Challenge 2021 is used as the baseline.} is used as the baseline method, upon which all the ablation experiments for text reading expert is conducted. First, it is observed that text reading expert greatly improves the performance on the ``Number'' type by over 6\%, where many questions are asked about reading numbers from OCR text, such as a bus number and a football player number. On the ``Other'' type, the performance can be improved by over 1\%. Answering many questions of this type requires the ability to reason with both visual and textual OCR information. Adding the separator between textual bounding boxes and continual pre-training on domain data can lead to further improvement, demonstrating the effectiveness of adapting the pre-trained StructuralLM for text-reading VQA. 

\paragraph{Clock Reading Expert}
The ablation study of clock reading expert is shown in Table~\ref{tab:clock}, where only the results on the ``Number'' and ``Overall'' types are given, because questions on reading clocks are only present in the ``Number'' type. Adding clock reading expert results in more than 4.5\% performance improvement on the ``Number'' type (from 59.93 to 62.65), which demonstrates the effectiveness of proposed ideas in the clock reading expert.
Specifically, the proposed regression loss is prone to provide a larger gradient when there is a bigger difference between the predicted time and the ground truth, which benefits prediction of the clock reader. Moreover, it can be observed that the self-supervised loss boosts the performance significantly, as the relationship prior constrains hour and minute branches both, which eliminates the confusion of hour and minute hands.

\begin{table}[t]
\caption{Ablation study of clock reading expert. } 
\centering
\begin{tabular}{c|ccc|c|cc}
\toprule
Clock Detector         & \multicolumn{4}{c|}{Clock Reader}                        & \multicolumn{2}{c}{VQA Test-dev} \\ \midrule
Detection(mAP)         & Cls& Regression Loss & Self-supervised Loss & Clock Accuracy & Number    & Overall   \\ \midrule
Baseline                     & --       & --       & --               & --         & 59.93        & 76.51          \\
\multirow{3}{*}{79.30} &\checkmark&          &                  & 72.5       & 62.52        & 76.79          \\
                       &\checkmark&\checkmark&                  & 73.0       & 62.59        & 76.80           \\
                       &\checkmark&\checkmark&\checkmark        & \textbf{74.7}       & \textbf{62.65}        & \textbf{76.81}        \\
\bottomrule
\end{tabular}
\label{tab:clock}
\end{table}

\begin{table}[t]
\centering
\caption{Ablation study of MoE on the VQA Test-dev.}\label{tab:moe_result}
\begin{tabular}{lcccc}
\toprule
           & Overall & Yes/No & Number & Other \\
\midrule
Visual Understanding Expert &  80.05    &   93.67     &   66.78     &  71.40     \\
\quad + Text Reading Expert (MoE) &  81.00  &   93.67  & 69.75     &  \bf 72.69     \\
\quad \quad + Clock Reading Expert (MoE)   &  \bf 81.27   & 93.67  & \bf 72.55    &  72.60  \\
\bottomrule
\end{tabular}
\end{table}




\paragraph{Mixture of Experts (MoE)}
The ablation study of MoE is shown in Table \ref{tab:moe_result}. With only visual understanding expert, the model gives a strong performance of accuracy 80.05 on the VQA Test-dev set. Adding text reading expert increases the overall performance by more than 1\%, which already achieves human parity of 80.83. Adding clock reading expert further boosts the performance to 81.27, where the performance on the ``Number'' type increases by more than 4\%. The gating network of MoE mimics human who is able to identify domain experts based on the nature of tasks. This knowledge-guided MoE framework can also be easily extended to incorporate more specialized experts for continual self-evolution.

\begin{figure}[b]
\includegraphics[width=\textwidth]{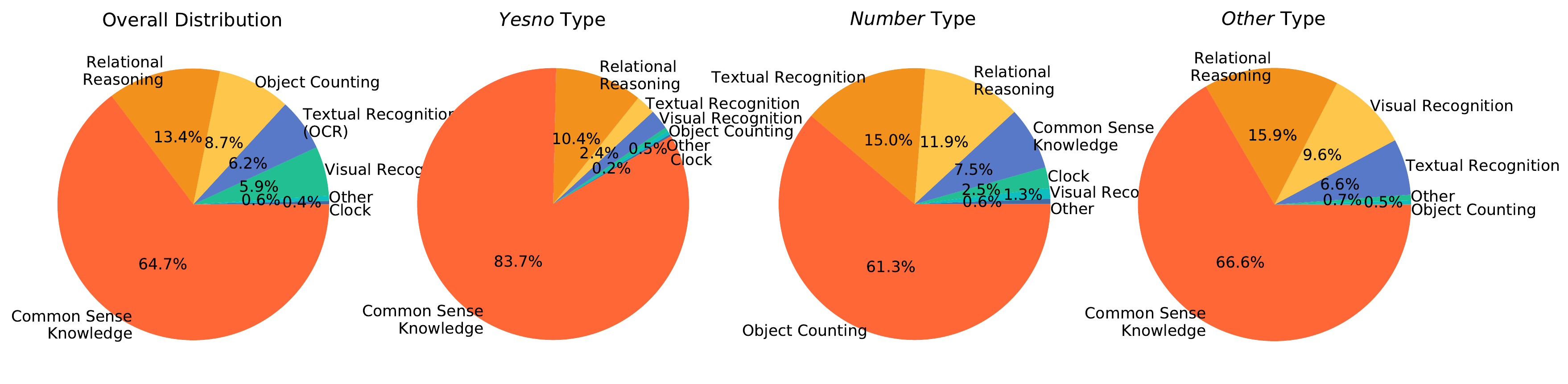}
\caption{The distribution of abilities in each answer type.}
\label{fig:vqa_answertype}
\end{figure}

\subsection{VQA Dataset Analysis}

\begin{figure}[t] 
    \centering
    \includegraphics[width=\textwidth]{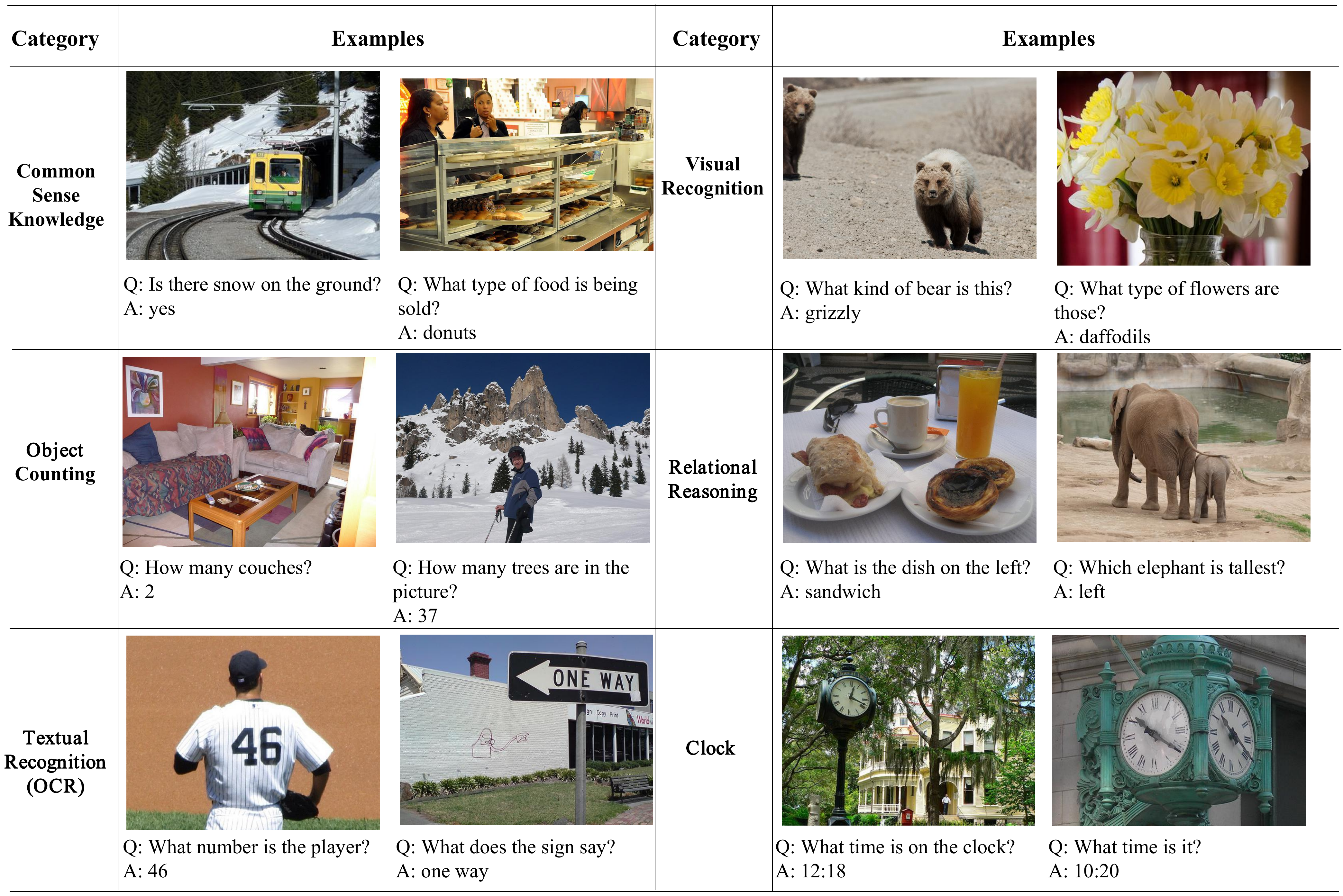}
    \caption{Representative examples from each category.}
    \label{fig:examples}
\end{figure}

This subsection provides more detailed analysis of our VQA results. To gain an understanding of types of questions and answers, 1000 examples are randomly sampled from the validation set for analysis.
The 1000 examples are classified into the categories listed below by manual examination based on the abilities required. The categorization is multi-label in the sense that every example is classified into all applicable categories. Figure~\ref{fig:vqa_answertype} provides an estimate of the proportion for each category. Commonsense Knowledge, Relational Reasoning and Object Counting are the top three categories in the overall distribution. Commonsense Knowledge accounts for over 80\% of the \emph{Yes/No} type. In the \emph{Number} type, Object Counting and Textual Recognition are the two most popular categories compared with the other two types. The type \emph{Other} has a similar distribution as that of the overall distribution. Figure \ref{fig:examples} presents representative examples from each category.

\begin{itemize}
    \item \textbf{Commonsense Knowledge} This category contains questions inquiring commonsense knowledge from our daily life, such as colors, weathers, food and furniture.
    \item \textbf{Visual Recognition} This category requires the ability to acquire specialized knowledge with visual recognition to answer questions in this category.
    \item \textbf{Relational Reasoning} This category requires understanding and reasoning over certain relationships of objects in an image, such as positional relationship, comparison relationship, etc.    
    \item \textbf{Textual Recognition (OCR)} This category requires the ability to recognize and utilize text together with the positions or visual information in an image (e.g., road signs, ads on a bus).
    \item \textbf{Object Counting} This category contains the examples that test the ability of counting objects in an image.
    \item \textbf{Clock Reading} This category contains the examples that test the ability of reading a clock.
    \item \textbf{Other} This category contains the questions that are ambiguous or cannot be answered based on given images.
\end{itemize}

\begin{table}[t]
\caption{The overall performance of \modelname and human on val split.}\label{tab:human_overall}
\setlength{\tabcolsep}{0.6mm}
\centering
\begin{tabular}{l|c|cccc}
\toprule
\multirow{2}{*}{} & Test-std & \multicolumn{4}{c}{Val}           \\
\cmidrule{2-6}
                  & Overall        & Overall & Yes/no & Number & Other \\
\midrule
VLP     &    \bf 81.26    &  79.54       & 92.47      & 70.63      & \bf 72.00     \\
Human                                 & 80.83          & 78.69   & \bf 94.87  & \bf 78.79  &  66.34\\
\bottomrule
\end{tabular}
\end{table}

\begin{table}[t]
\centering
\setlength{\tabcolsep}{0.6mm}
\caption{The performance of \modelname and human by category.}\label{tab:human_category}
\begin{tabular}{c|ccccccc}
\toprule
      & \tabincell{c}{Commonsense \\Knowledge} & \tabincell{c}{Relational \\Reasoning} & Object Counting & Visual Recognition & \tabincell{c}{Textual Recognition \\ (OCR)} & Clock Reading & Other \\
      \cmidrule{2-8}
      & 767 & 159  & 103 & 70  & 74 & 7  & 5  \\
\midrule
VLP   & \bf 83.60   & \bf 71.19 & 77.76  & \bf 68.14     & 52.03 & \bf 86.00 & \bf  70.00 \\
Human & 80.04                  & 70.20     & \bf  81.29           & 59.76              & \bf 76.62                     & 60.66 & 49.52 \\
\bottomrule
\end{tabular}
\end{table}

\subsection{\modelname vs. Human}
%


A comparative study of \modelname and human on visual question answering has been conducted. Table \ref{tab:human_overall} and Table \ref{tab:human_category} show the overall and per-category performance of \modelname and human on the val split, respectively, from which there are the following observations: 
(i) \modelname outperforms human annotators on the two largest categories, Commonsense Knowledge and Relational Reasoning. It shows \modelname's superiority of identifying common scene objects in daily life and leveraging commonsense knowledge such as colors and weathers. This result also demonstrates the power of \modelname in reasoning over relative positions, such as the left sign on a wall, to answer a spatial reasoning question. Besides, it is surprising that \modelname can reason over simple comparison, such as which object is the tallest.
(ii) The questions in the Object Counting category seem rather difficult for \modelname to answer. \modelname is found to be good at counting a small number (<10) of objects. It would give an incorrect count when encountering a large number of small objects and/or requiring reasoning over them.
(iii) \modelname significantly surpasses human performance on Visual Recognition which requires specialized knowledge. It is expected that \modelname, as a machine learner trained with large data, is skilled in memorizing specialized/professional knowledge with visual recognition, compared with non-professional human annotators.
(iv) \modelname is more capable of reading time shown in a clock than human, as demonstrated by the result of Clock Reading. On text reading, however, there is still a big gap between \modelname and human in recognizing and understanding text in an image, as shown by the result of Textual Recognition. Some research progress has been made on text-reading VQA tasks, such as TextVQA~\citep{textvqa}.



\begin{figure}[t] 
    \centering
    \includegraphics[width=\textwidth]{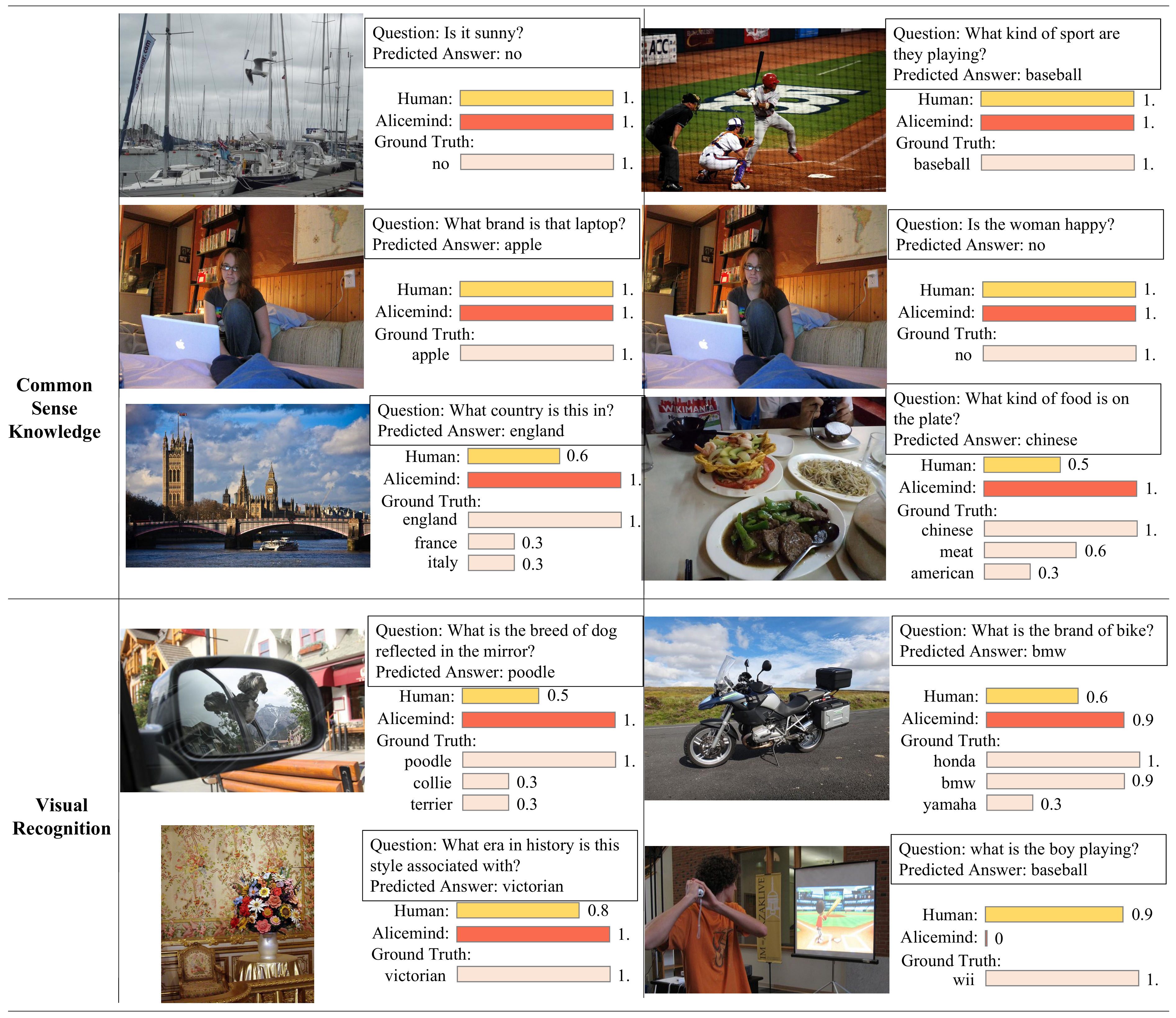}
    \caption{Case Study for Commonsense Knowledge and Visual Recognition. The scores of Human and \modelname are calculated with Equation~\ref{eq:evaluation}. Ground Truth gives the top three annotations.}
    \label{fig:case_study_1}
\end{figure}

\begin{figure}[t] 
    \centering
    \includegraphics[width=\textwidth]{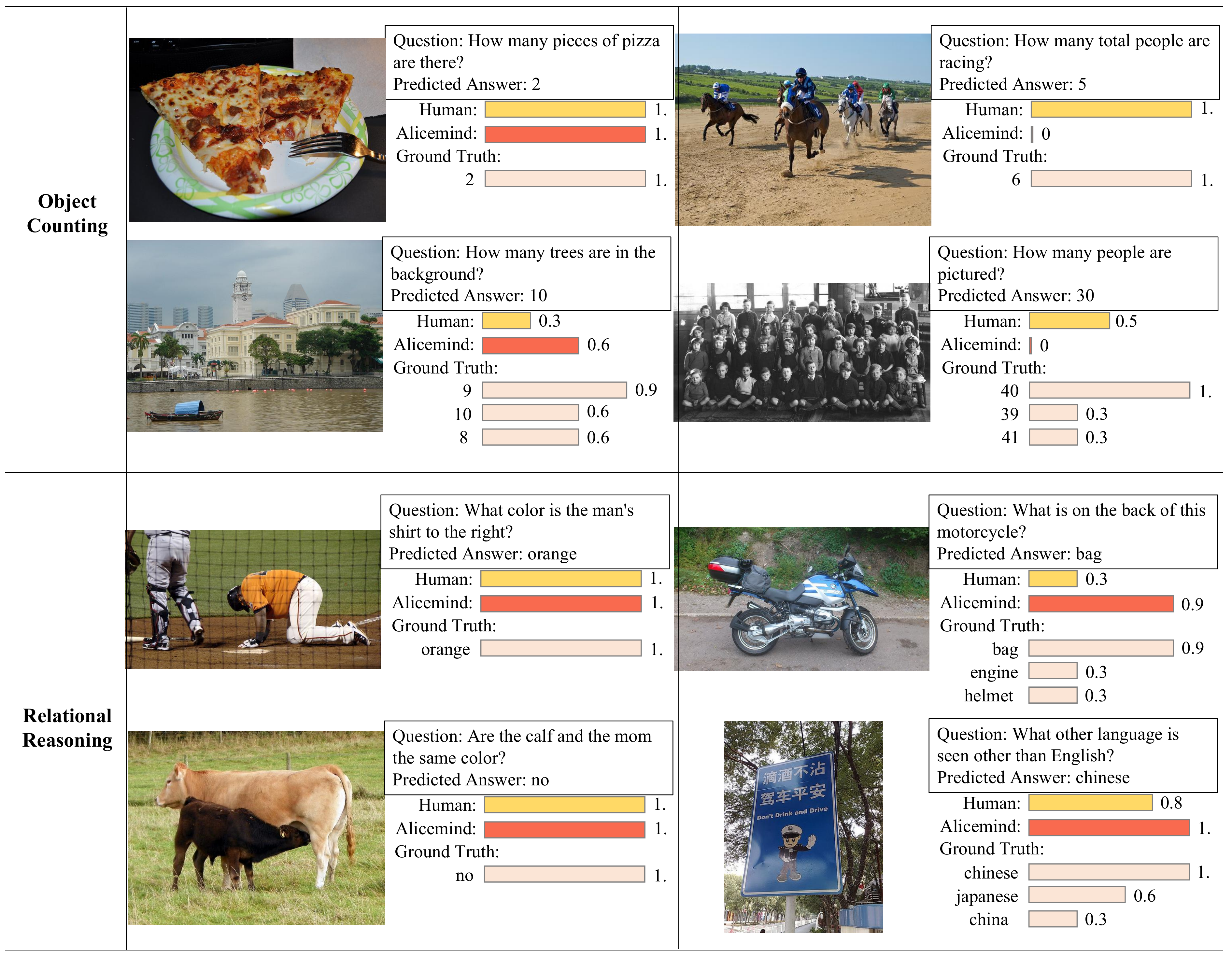}
    \caption{Case Study for Object Counting and Relational Reasoning.}
    \label{fig:case_study_2}
\end{figure}

\begin{figure}[t] 
    \centering
    \includegraphics[width=\textwidth]{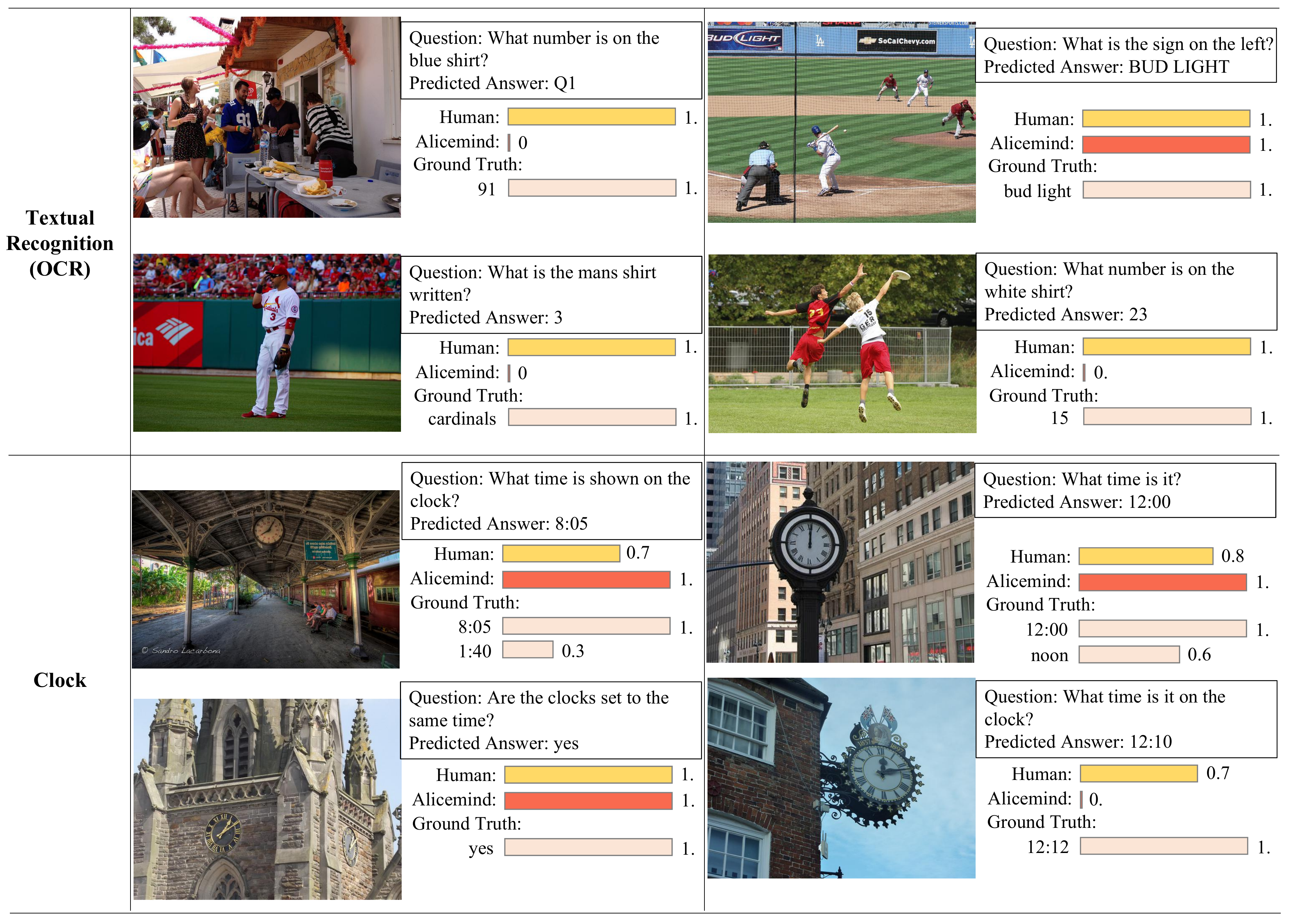}
    \caption{Case Study for Textural Recognition (OCR) and Clock Reading.}
    \label{fig:case_study_3}
\end{figure}

Figure~\ref{fig:case_study_1}, Figure~\ref{fig:case_study_2} and Figure~\ref{fig:case_study_3} present \modelname's predictions together with ground truth on each category for case study. In particular, a couple of representative examples are listed for each category, each containing a question, an image and the answer predicted by \modelname. The scores of human annotators and \modelname, as well as the top three ground-truth annotations are also given for comparison. The scores of Human and \modelname are calculated based on Equation~\ref{eq:evaluation}. These examples are studied by category as follows:

\paragraph{Commonsense Knowledge}
\modelname is knowledgeable in many aspects of daily life. As shown in Figure~\ref{fig:case_study_1}, \modelname is able to tell not only weather and the sentiments of people, but also classic sports and electronic products as an ordinary person does. Also, it is skilled in geography and understands the food around the world. For example, \modelname recognizes the small English flag and Big Ben in the image, by which the country is identified as England. As another example, based on the rice and dishes from the image, Chinese food can be identified by \modelname and people familiar with Chinese cuisine. The other people may not tell cuisine of the food .
Our experiments show that \modelname trained on adequate data can capture the commonsense knowledge (the largest category in the VQA dataset) in our daily life.

\paragraph{Visual Recognition}
Except for Clock Reading, \modelname shows much better performance than an ordinary person does in this category. As shown in Figure~\ref{fig:case_study_1}, it is relatively easy for an AI model trained adequately to memorize specialized knowledge, while rather difficult for people unfamiliar with the specific domain. For example, \modelname can better identify the specific categories of the animals, such as dog and bird, and the historical style of the furniture, which requires specialized knowledge. By locating and recognizing the barely visible logo from the motor bike, \modelname correctly recognizes its brand, while human may miss the details in the image and give an answer based on their best guesses. On the other hand, \modelname has a slim chance of being fooled by the activity present in the image. It incorrectly identifies that the boy is playing baseball based on his motion, while it is actually a Wii game. As a result, by incorporating relevant specialized knowledge and visual recognition capability, \modelname can outperform human by a large margin in this category.

\paragraph{Object Counting}
As shown in Figure~\ref{fig:case_study_2}, \modelname achieves human parity when counting a small number of objects, but fails in more complicated cases where there are occlusions or a great quantity of objects. It surprises us that \modelname can give a count very close to the correct answer in the example of counting trees. However, the object counting ability is still quite limited compared with ordinary people. One reason may lie in that the visual detection is too weak to detect all the objects in an image when the number of objects is large. There are few cases with more than 10 objects in the training set, and thus the model is not fully trained with sufficient data. This is shown in the third example where \modelname fails to identify all people from the image. Another possible reason is that the object detector is difficult to count in the presence of occlusion. The second example shows that \modelname counts the racing people incorrectly due to the occluded person.

\paragraph{Relational Reasoning}
Figure~\ref{fig:case_study_2} shows that \modelname has the abilities to reason over the relationship of positions, comparison and exclusion. It is observed that \modelname may be more capable than human in precise position identification and knowledge reasoning for relational reasoning questions. Specifically, 1) \emph{position}: the first two examples show the power of \modelname in distinguishing the positions of the left-right and front-back, and conducting one-step reasoning over the positional relationship; 2) \emph{comparison}: the third example demonstrates that \modelname can even compare the colors of two objects, which is a simple one-step reasoning over the comparison between attributes of two objects; 3) \emph{exclusion}: the last example shows that \modelname is able to identify the exclusion relationship, and reason over it with commonsense knowledge.

\paragraph{Textual Recognition (OCR)}
As shown in Figure~\ref{fig:case_study_3}, the text reading expert (StructuralLM) is able to identify text and layout in simple cases. In the first example, the model correctly answers with the words displayed on the man's shirt. In addition, StructuralLM is capable of learning the interactions between text and layout, which makes StructuralLM aware of location of text present in an image. This is shown in the second example, where the model predicts the answer correctly when asked about the sign on the left. However, the model fails in the two cases: 1) OCR errors; 2) answering complex questions which requires visual feature and reasoning abilities. As shown in the third example, when asked the words on the man's shirt, the model can predict only ``3'' because the OCR tool cannot recognize the word ``cardinals''. In the fourth example, given the question about the number present on the white shirt, the model answers incorrectly due to the lack of visual feature of colors and the reasoning ability. Currently, the text reading expert utilizes only the layout and textual information to answer a text-reading question, without leveraging visual signals. There is an urgent need for deep interaction between visual information and OCR textual information in images, which is left for future work.

\paragraph{Clock Reading}
As shown in Figure~\ref{fig:case_study_3}, the clock reading expert is able to read the clock time accurately at a five-minute level. One important problem to be addressed is distinguishing the hour hand (generally shorter) from the minute hand (generally longer). The clock reading expert is trained well on this objective. Therefore, in the first example, the model predicts the correct time ``8:05'', while some human annotators misread the hour hand and minute hand, and thus give the wrong time reading ``1:40''. In the second example, the clock reading expert can tell the time accurately even when the hour hand and minute hand overlap. There also exist limitations for the current clock reading expert, that it can't tell more accurate time at a minute level. In the fourth example, the model only recognizes the time is about 12:10, but cannot tell the exact time of 12:12. The reason comes from casting the problem as detection and then classification, of which the clock training data is not adequate to support training of 1-minute level clock reading.

%% file: rel_work.tex
\subsection{Visual Question Answering}
First released in 2016, the VQA task and research on the dataset have been developing for more than five years. A large number of new methods have been proposed to push the limit of this task. Existing approaches on the VQA task mostly put efforts on four key aspects: 1) better visual feature representation, 2) effective cross-modal fusion, 3) large-scale vision-and-language pre-training, and 4) task-specific expert optimization. Region-based visual features have long been treated as the \emph{de facto} standard for vision and language tasks such as VQA, where visual semantics and salient image objects are captured with bottom-up attention~\citep{anderson2018bottom}. The proposed bottom-up attention method with region features won the VQA Challenge of 2017, which has been largely adopted since then. Recently, Li et al.~\citep{zhang2021vinvl} has further pushed the limit of the region-based features by pre-training on extremely large object detection datasets. On the other hand, \citep{jiang2020defense} studied the key factors that contributed to the effectiveness of existing bottom-up attention features, and found that the grid-based features from convolutional neural networks can yield comparable or even better results on the VQA task. It won the VQA Challenge of 2020.
More recently, inspired by the popular vision Transformer~\citep{dosovitskiy2020image}, some pioneer work such as ViLT~\citep{kim2021vilt} began to study the use of patch-based visual feature for its efficiency.

Cross-modal fusion or interaction has always been treated as one of the most important challenges in cross-modality research. At first, the cross-modal fusion on VQA is simply element-wise product of textual and visual representation~\citep{vqa}. Then, it gradually improves from a linear model~\citep{vqa} to bilinear pooling ones~\citep{fukui2016multimodal,yu2017multi}, and attention-based fusion ones~\citep{lu2016hierarchical,yang2016stacked,yu2019deep}, where questions and image features are fully interacted with each other. Recently, with the popularity of \emph{pre-train and fine-tune} paradigm such as BERT~\citep{devlin2018bert} in NLP, large-scale vision-and-language pre-training~\citep{tan2019lxmert,su2019vl,li2020unicoder,chen2020uniter,li2020oscar,huang2020pixel,yu2021ernie} has been used to better align the vision-language representations. It has established the new state-of-the-art performance on the VQA task since 2020, by pre-training on a large amount of unlabeled image-text pairs.

To further improve the VQA performance, some studies analyze the weakness of existing models on the VQA task, and address it with specific expert optimization. For example, MoVie~\citep{nguyen2020movie} revisited modulated convolutions for visual counting, which is more efficient and advances the state-of-the-art on counting-specific VQA tasks. \citep{wu2016ask} propose a knowledge-enhanced VQA method to combine an internal representation of image content with information extracted from a general knowledge base to answer a broad range of image-based questions.

\subsection{Visual Features for Vision-and-Language tasks}
Feature representations for vision have played a key role in the advancement of vision and language tasks. Different kinds of features can capture diverse data characteristics, which complements with each other. In most of the vision-and-language tasks, visual feature is the bottleneck of existing multi-modal models and will set the upper bound for the final performance. To capture both the local and global information in the image, this work takes a comprehensive visual feature understanding on three kinds of typical features.

\paragraph{Region Feature} 
Popularized as bottom-up attention~\citep{anderson2018bottom}, region-based visual features have been treated as the de facto standard for vision and language tasks and achieved dominant performance in tasks such as visual question answering (VQA) and image captioning~\citep{chen2015microsoft}. It uses pre-trained object detectors~\citep{ren2015faster} to identify salient regions based on the vision input. A region proposal network (RPN) is first used to propose regions of interest (RoI) based on the grid features pooled from the CNN backbone. Then non-maximum suppression (NMS) is used to select a small collection of proper RoIs, and RoI head is used to extract region feature for each selected RoI. As a result, images are represented by a collection of region-based features. Most current VLP approaches such as LXMERT~\citep{tan2019lxmert}, UNITER~\citep{chen2020uniter}, ERNIE-ViL~\citep{yu2021ernie} and OSCAR~\citep{li2020oscar} adopt the region feature-based VLP paradigm. Recently, Li et al.~\citep{zhang2021vinvl} have proposed an extreme large version of the region-based object detector, which is pretrained on four object detection datasets, including MS COCO~\citep{lin2014microsoft}, Visual Genome~\citep{krishna2017visual}, OpenImages~\citep{kuznetsova2020open}, and Object365~\citep{shao2019objects365}. It greatly improved the performance of VLP models, creating new state-of-the-art results on seven public benchmarks. 

\paragraph{Grid Feature}
Besides the local region-based features, the grid-like feature map from convolutional neural networks such as ResNet~\citep{he2016deep} can also be used as visual features for visual-and-language tasks. One of the largest advantages of grid feature is that it removes all the time-consuming region-related steps, which can support more flexible architecture design of end-to-end vision-and-language models. Successful use of grid feature was first proposed by GridFeat~\citep{jiang2020defense} on VQA and image captioning task. It adopts the same pre-training settings as in the region-based method, discarding all the region-related steps from the detector and using only the grid convolutional features during inference. In this way, it greatly reduces the inference time of visual encoder and obtains comparable or even better accuracy on VQA task. Grid-VLP~\citep{yan2021gridvlp} further uses grid features for vision-language pre-training, and by using only in-domain datasets, Grid-VLP outperforms most state-of-the-art region-based VLP methods. Pixel-BERT~\citep{huang2020pixel} is the first end-to-end grid-based VLP method, which uses a ResNet image encoder and jointly optimizes both ResNet image encoder and cross-modality Transformer in an end-to-end manner. E2E-VLP~\citep{xu2021e2e} builds upon the detection transformer (DETR)~\citep{carion2020end}, and proposes an end-to-end grid-based VLP model for both V+L understanding and generation. It enhances learning of the pre-trained model by capturing more semantic-related visual representation with object detection and image caption pretext tasks.

\paragraph{Patch Feature}
Patch feature gains its popularity with the rise of Vision Transformer (ViT)~\citep{dosovitskiy2020image}, which firstly splits an image into fixed-size patches, then uses a simple linear projection of a patch before feeding them into transformers. Vision Transformer (ViT)~\citep{dosovitskiy2020image} is the first successful use of self-attention architecture to replace the convolutional neural network architecture in computer vision, after which a series of related work~\citep{liu2021swin, touvron2021training, chen2021pre} has been proposed to promote the development of this new direction. Swin Transformer~\citep{liu2021swin} further builds the model on multi-scale feature maps and uses a shifted windowing scheme on image patches, which makes it compatible with a broad range of vision tasks such as image classification and object detection. Most of the work in this ViT series adopts the image patch feature so as to reserve high efficiency in self attention-based Transformer modeling. CLIP~\citep{radford2021learning} proposes a simple contrastive pre-training method to learn the visual models based on a dataset of 400 million image-text pairs, which obtains superior zero-shot performance on various existing computer vision datasets. Recently, ViLT~\citep{kim2021vilt} has proposed the pioneering work of vision-language pre-training on image patch feature, and achieved up to tens of times faster than previous VLP models with a convolution-free scheme. 

\subsection{Vision-and-Language Pre-training}
Inspired by the breakthrough of language pre-training in NLP field, the  research  community begins to pay more attention to vision-language pre-training on large-scale image-text pairs, which also shows great effectiveness and achieves state-of-the-art performance across a variety of vision-language (VL) tasks~\citep{antol2015vqa,suhr2018corpus,hudson2019gqa}. 

Existing approaches to VLP~\citep{tan2019lxmert,su2019vl,li2020unicoder,chen2020uniter,li2020oscar,huang2020pixel,yu2021ernie} mainly take a two-step training pipeline, which first extracts semantic visual features by specific object detector and then learns a cross-modal pre-training model to align text and visual features. Current research about this topic can be roughly divided into two lines. The first line adopts a single-stream transformer architecture to model both image and text representations in a unified semantic space such as VLBERT~\citep{su2019vl}, UNITER~\citep{chen2020uniter} and OSCAR~\citep{li2020oscar}. In contrast, the other line uses a two-stream Transformer architecture that first encodes the image and text modalities separately, and then fuses the cross-modal representations with another Transformer network, such as LXMERT~\citep{tan2019lxmert} and ERNIE-ViL~\citep{yu2021ernie}. Recently, VinVL~\citep{zhang2021vinvl} 
pre-trained a large-scale object-attribute detection model with much larger amounts of data on four public object detection datasets for extracting better region feature, and creating new state-of-the-art results on seven public benchmarks.  In addition to image-text pairs, UNIMO~\citep{li2020unimo} also employed large scale of free text corpus and image collections for enhancing the cross-modal learning. These methods rely heavily on a task-specific bounding box (or region) based object detector, which impose unnecessary constraints on model designs and limit potential applications of existing vision and language systems. Therefore, PixelBERT~\citep{huang2020pixel} and E2E-VLP~\citep{xu2021e2e} further proposed end-to-end VLP method, by jointly learning both the visual encoder and cross-modal Transformer simultaneously. Besides, the object detector is removed from the whole process, and the VLP model directly conducts on the grid-based feature map from convolutional visual encoder. To further improve the training and inference speed of VLP model, ViLT~\citep{kim2021vilt} removed both the region supervision and convolutional visual encoder, and conducted vision-language pre-training directly on image patch feature with linear projection.

%% file: conclusion.tex


This paper describes our new research work on improving the full pipeline of the VQA task, which has achieved human parity on this challenging task for the first time. The key to the breakthrough lies in three aspects: 1) more comprehensive textual and visual feature representation with pre-trained technologies, 2) more effective cross-modal interaction with learning to attend, and 3) more knowledge-guided optimization with mixture of expert modules. It demonstrates the power of an AI model to achieve human parity on the challenging cross-modal understanding task under a closed-set scenario, which requires AI to have the ability to understand both visual and textual information. This makes it possible to further conduct higher-order cognition and commonsense reasoning intelligence.

Despite the success, the current AI technology on V\&L understanding still has notable limitations, and bridging the gap between machine intelligence and real human intelligence still has a long way to go. For the VQA task, there still exist certain weaknesses for our model: 1) object counting is still a very difficult problem for the current VLP model, especially the case when a large number of different objects with a small or tiny size exist. Besides, some objects can even overlap with each other, which introduces more complexity; 2) composite reading of both OCR text and visual content is still challenging for the current VLP model, while it is relatively easy for human to answer; 3) the current model can only work in a closed-set scenario, which is not applicable to the open-domain or unseen set. In the future, we hope to witness more breakthrough on open-set learning and more intelligent AI models, which can evolve itself by acquiring and reasoning about new knowledge.

%% file: main.bbl
\begin{thebibliography}{75}
\providecommand{\natexlab}[1]{#1}
\providecommand{\url}[1]{\texttt{#1}}
\expandafter\ifx\csname urlstyle\endcsname\relax
  \providecommand{\doi}[1]{doi: #1}\else
  \providecommand{\doi}{doi: \begingroup \urlstyle{rm}\Url}\fi

\bibitem[Silver et~al.(2016)Silver, Huang, Maddison, Guez, Sifre, van~den
  Driessche, Schrittwieser, Antonoglou, Panneershelvam, Lanctot, Dieleman,
  Grewe, Nham, Kalchbrenner, Sutskever, Lillicrap, Leach, Kavukcuoglu, Graepel,
  and Hassabis]{alphago}
David Silver, Aja Huang, Christopher~J. Maddison, Arthur Guez, Laurent Sifre,
  George van~den Driessche, Julian Schrittwieser, Ioannis Antonoglou, Veda
  Panneershelvam, Marc Lanctot, Sander Dieleman, Dominik Grewe, John Nham, Nal
  Kalchbrenner, Ilya Sutskever, Timothy Lillicrap, Madeleine Leach, Koray
  Kavukcuoglu, Thore Graepel, and Demis Hassabis.
\newblock Mastering the game of go with deep neural networks and tree search.
\newblock \emph{Nature}, 529:\penalty0 484--503, 2016.

\bibitem[Deng et~al.(2009{\natexlab{a}})Deng, Dong, Socher, Li, Li, and
  Fei-Fei]{imagenet}
Jia Deng, Wei Dong, Richard Socher, Li-Jia Li, Kai Li, and Li~Fei-Fei.
\newblock Imagenet: A large-scale hierarchical image database.
\newblock In \emph{2009 IEEE conference on computer vision and pattern
  recognition}, pages 248--255. Ieee, 2009{\natexlab{a}}.

\bibitem[Wang et~al.(2018)Wang, Singh, Michael, Hill, Levy, and Bowman]{glue}
Alex Wang, Amanpreet Singh, Julian Michael, Felix Hill, Omer Levy, and Samuel
  Bowman.
\newblock {GLUE}: A multi-task benchmark and analysis platform for natural
  language understanding.
\newblock In \emph{Proceedings of the 2018 {EMNLP} Workshop {B}lackbox{NLP}:
  Analyzing and Interpreting Neural Networks for {NLP}}, pages 353--355,
  Brussels, Belgium, November 2018. Association for Computational Linguistics.

\bibitem[Sharif~Razavian et~al.(2014)Sharif~Razavian, Azizpour, Sullivan, and
  Carlsson]{razavian2014}
Ali Sharif~Razavian, Hossein Azizpour, Josephine Sullivan, and Stefan Carlsson.
\newblock Cnn features off-the-shelf: an astounding baseline for recognition.
\newblock In \emph{Proceedings of the IEEE Conference on Computer Vision and
  Pattern Recognition Workshops}, pages 806--813, 2014.

\bibitem[Devlin et~al.(2018)Devlin, Chang, Lee, and Toutanova]{devlin2018bert}
Jacob Devlin, Ming-Wei Chang, Kenton Lee, and Kristina Toutanova.
\newblock Bert: Pre-training of deep bidirectional transformers for language
  understanding.
\newblock \emph{arXiv preprint arXiv:1810.04805}, 2018.

\bibitem[Agrawal et~al.(2017)Agrawal, Lu, Antol, Mitchell, Zitnick, Parikh, and
  Batra]{vqa}
Aishwarya Agrawal, Jiasen Lu, Stanislaw Antol, Margaret Mitchell, C.~Lawrence
  Zitnick, Devi Parikh, and Dhruv Batra.
\newblock Vqa: Visual question answering.
\newblock \emph{Int. J. Comput. Vision}, 123\penalty0 (1):\penalty0 4–31, May
  2017.

\bibitem[Su et~al.(2019)Su, Zhu, Cao, Li, Lu, Wei, and Dai]{su2019vl}
Weijie Su, Xizhou Zhu, Yue Cao, Bin Li, Lewei Lu, Furu Wei, and Jifeng Dai.
\newblock Vl-bert: Pre-training of generic visual-linguistic representations.
\newblock \emph{arXiv preprint arXiv:1908.08530}, 2019.

\bibitem[Chen et~al.(2020)Chen, Li, Yu, El~Kholy, Ahmed, Gan, Cheng, and
  Liu]{chen2020uniter}
Yen-Chun Chen, Linjie Li, Licheng Yu, Ahmed El~Kholy, Faisal Ahmed, Zhe Gan,
  Yu~Cheng, and Jingjing Liu.
\newblock Uniter: Universal image-text representation learning.
\newblock In \emph{European conference on computer vision}, pages 104--120.
  Springer, 2020.

\bibitem[Tan and Bansal(2019)]{tan2019lxmert}
Hao Tan and Mohit Bansal.
\newblock Lxmert: Learning cross-modality encoder representations from
  transformers.
\newblock \emph{arXiv preprint arXiv:1908.07490}, 2019.

\bibitem[Yu et~al.(2021)Yu, Tang, Yin, Sun, Tian, Wu, and Wang]{yu2021ernie}
Fei Yu, Jiji Tang, Weichong Yin, Yu~Sun, Hao Tian, Hua Wu, and Haifeng Wang.
\newblock Ernie-vil: Knowledge enhanced vision-language representations through
  scene graphs.
\newblock In \emph{Proceedings of the AAAI Conference on Artificial
  Intelligence}, volume~35, pages 3208--3216, 2021.

\bibitem[Anderson et~al.(2018)Anderson, He, Buehler, Teney, Johnson, Gould, and
  Zhang]{anderson2018bottom}
Peter Anderson, Xiaodong He, Chris Buehler, Damien Teney, Mark Johnson, Stephen
  Gould, and Lei Zhang.
\newblock Bottom-up and top-down attention for image captioning and visual
  question answering.
\newblock In \emph{Proceedings of the IEEE conference on computer vision and
  pattern recognition}, pages 6077--6086, 2018.

\bibitem[Ren et~al.(2015)Ren, He, Girshick, and Sun]{ren2015faster}
Shaoqing Ren, Kaiming He, Ross Girshick, and Jian Sun.
\newblock Faster r-cnn: Towards real-time object detection with region proposal
  networks.
\newblock \emph{Advances in neural information processing systems},
  28:\penalty0 91--99, 2015.

\bibitem[Krishna et~al.(2017)Krishna, Zhu, Groth, Johnson, Hata, Kravitz, Chen,
  Kalantidis, Li, Shamma, et~al.]{krishna2017visual}
Ranjay Krishna, Yuke Zhu, Oliver Groth, Justin Johnson, Kenji Hata, Joshua
  Kravitz, Stephanie Chen, Yannis Kalantidis, Li-Jia Li, David~A Shamma, et~al.
\newblock Visual genome: Connecting language and vision using crowdsourced
  dense image annotations.
\newblock \emph{International journal of computer vision}, 123\penalty0
  (1):\penalty0 32--73, 2017.

\bibitem[Zhang et~al.(2021)Zhang, Li, Hu, Yang, Zhang, Wang, Choi, and
  Gao]{zhang2021vinvl}
Pengchuan Zhang, Xiujun Li, Xiaowei Hu, Jianwei Yang, Lei Zhang, Lijuan Wang,
  Yejin Choi, and Jianfeng Gao.
\newblock Vinvl: Revisiting visual representations in vision-language models.
\newblock In \emph{Proceedings of the IEEE/CVF Conference on Computer Vision
  and Pattern Recognition}, pages 5579--5588, 2021.

\bibitem[Li et~al.(2021{\natexlab{a}})Li, Yan, Xu, Luo, Wang, Bi, and
  Huang]{li2021semvlp}
Chenliang Li, Ming Yan, Haiyang Xu, Fuli Luo, Wei Wang, Bin Bi, and Songfang
  Huang.
\newblock Semvlp: Vision-language pre-training by aligning semantics at
  multiple levels.
\newblock \emph{arXiv preprint arXiv:2103.07829}, 2021{\natexlab{a}}.

\bibitem[Huang et~al.(2020)Huang, Zeng, Liu, Fu, and Fu]{huang2020pixel}
Zhicheng Huang, Zhaoyang Zeng, Bei Liu, Dongmei Fu, and Jianlong Fu.
\newblock Pixel-bert: Aligning image pixels with text by deep multi-modal
  transformers.
\newblock \emph{arXiv preprint arXiv:2004.00849}, 2020.

\bibitem[Xu et~al.(2021)Xu, Yan, Li, Bi, Huang, Xiao, and Huang]{xu2021e2e}
Haiyang Xu, Ming Yan, Chenliang Li, Bin Bi, Songfang Huang, Wenming Xiao, and
  Fei Huang.
\newblock E2e-vlp: End-to-end vision-language pre-training enhanced by visual
  learning.
\newblock \emph{arXiv preprint arXiv:2106.01804}, 2021.

\bibitem[Yan et~al.(2021)Yan, Xu, Li, Bi, Tian, Gui, and Wang]{yan2021gridvlp}
Ming Yan, Haiyang Xu, Chenliang Li, Bin Bi, Junfeng Tian, Min Gui, and Wei
  Wang.
\newblock Grid-vlp: Revisiting grid features for vision-language pre-training,
  2021.

\bibitem[Jiang et~al.(2020)Jiang, Misra, Rohrbach, Learned-Miller, and
  Chen]{jiang2020defense}
Huaizu Jiang, Ishan Misra, Marcus Rohrbach, Erik Learned-Miller, and Xinlei
  Chen.
\newblock In defense of grid features for visual question answering.
\newblock In \emph{Proceedings of the IEEE/CVF Conference on Computer Vision
  and Pattern Recognition}, pages 10267--10276, 2020.

\bibitem[He et~al.(2016)He, Zhang, Ren, and Sun]{he2016deep}
Kaiming He, Xiangyu Zhang, Shaoqing Ren, and Jian Sun.
\newblock Deep residual learning for image recognition.
\newblock In \emph{Proceedings of the IEEE conference on computer vision and
  pattern recognition}, pages 770--778, 2016.

\bibitem[Deng et~al.(2009{\natexlab{b}})Deng, Dong, Socher, Li, Li, and
  Fei-Fei]{deng2009imagenet}
Jia Deng, Wei Dong, Richard Socher, Li-Jia Li, Kai Li, and Li~Fei-Fei.
\newblock Imagenet: A large-scale hierarchical image database.
\newblock In \emph{2009 IEEE conference on computer vision and pattern
  recognition}, pages 248--255. Ieee, 2009{\natexlab{b}}.

\bibitem[Radford et~al.(2021)Radford, Kim, Hallacy, Ramesh, Goh, Agarwal,
  Sastry, Askell, Mishkin, Clark, et~al.]{radford2021learning}
Alec Radford, Jong~Wook Kim, Chris Hallacy, Aditya Ramesh, Gabriel Goh,
  Sandhini Agarwal, Girish Sastry, Amanda Askell, Pamela Mishkin, Jack Clark,
  et~al.
\newblock Learning transferable visual models from natural language
  supervision.
\newblock \emph{arXiv preprint arXiv:2103.00020}, 2021.

\bibitem[Jia et~al.(2021)Jia, Yang, Xia, Chen, Parekh, Pham, Le, Sung, Li, and
  Duerig]{jia2021scaling}
Chao Jia, Yinfei Yang, Ye~Xia, Yi-Ting Chen, Zarana Parekh, Hieu Pham, Quoc~V
  Le, Yunhsuan Sung, Zhen Li, and Tom Duerig.
\newblock Scaling up visual and vision-language representation learning with
  noisy text supervision.
\newblock \emph{arXiv preprint arXiv:2102.05918}, 2021.

\bibitem[Desai and Johnson(2021)]{desai2021virtex}
Karan Desai and Justin Johnson.
\newblock Virtex: Learning visual representations from textual annotations.
\newblock In \emph{Proceedings of the IEEE/CVF Conference on Computer Vision
  and Pattern Recognition}, pages 11162--11173, 2021.

\bibitem[Dosovitskiy et~al.(2020)Dosovitskiy, Beyer, Kolesnikov, Weissenborn,
  Zhai, Unterthiner, Dehghani, Minderer, Heigold, Gelly,
  et~al.]{dosovitskiy2020image}
Alexey Dosovitskiy, Lucas Beyer, Alexander Kolesnikov, Dirk Weissenborn,
  Xiaohua Zhai, Thomas Unterthiner, Mostafa Dehghani, Matthias Minderer, Georg
  Heigold, Sylvain Gelly, et~al.
\newblock An image is worth 16x16 words: Transformers for image recognition at
  scale.
\newblock In \emph{International Conference on Learning Representations}, 2020.

\bibitem[Liu et~al.(2021)Liu, Lin, Cao, Hu, Wei, Zhang, Lin, and
  Guo]{liu2021swin}
Ze~Liu, Yutong Lin, Yue Cao, Han Hu, Yixuan Wei, Zheng Zhang, Stephen Lin, and
  Baining Guo.
\newblock Swin transformer: Hierarchical vision transformer using shifted
  windows.
\newblock \emph{arXiv preprint arXiv:2103.14030}, 2021.

\bibitem[Touvron et~al.(2021)Touvron, Cord, Douze, Massa, Sablayrolles, and
  J{\'e}gou]{touvron2021training}
Hugo Touvron, Matthieu Cord, Matthijs Douze, Francisco Massa, Alexandre
  Sablayrolles, and Herv{\'e} J{\'e}gou.
\newblock Training data-efficient image transformers \& distillation through
  attention.
\newblock In \emph{International Conference on Machine Learning}, pages
  10347--10357. PMLR, 2021.

\bibitem[Chen et~al.(2021)Chen, Wang, Guo, Xu, Deng, Liu, Ma, Xu, Xu, and
  Gao]{chen2021pre}
Hanting Chen, Yunhe Wang, Tianyu Guo, Chang Xu, Yiping Deng, Zhenhua Liu, Siwei
  Ma, Chunjing Xu, Chao Xu, and Wen Gao.
\newblock Pre-trained image processing transformer.
\newblock In \emph{Proceedings of the IEEE/CVF Conference on Computer Vision
  and Pattern Recognition}, pages 12299--12310, 2021.

\bibitem[Kim et~al.(2021)Kim, Son, and Kim]{kim2021vilt}
Wonjae Kim, Bokyung Son, and Ildoo Kim.
\newblock Vilt: Vision-and-language transformer without convolution or region
  supervision.
\newblock \emph{arXiv preprint arXiv:2102.03334}, 2021.

\bibitem[Changpinyo et~al.(2021)Changpinyo, Sharma, Ding, and
  Soricut]{changpinyo2021conceptual}
Soravit Changpinyo, Piyush Sharma, Nan Ding, and Radu Soricut.
\newblock Conceptual 12m: Pushing web-scale image-text pre-training to
  recognize long-tail visual concepts.
\newblock In \emph{Proceedings of the IEEE/CVF Conference on Computer Vision
  and Pattern Recognition}, pages 3558--3568, 2021.

\bibitem[Liu et~al.(2019)Liu, Ott, Goyal, Du, Joshi, Chen, Levy, Lewis,
  Zettlemoyer, and Stoyanov]{liu2019roberta}
Yinhan Liu, Myle Ott, Naman Goyal, Jingfei Du, Mandar Joshi, Danqi Chen, Omer
  Levy, Mike Lewis, Luke Zettlemoyer, and Veselin Stoyanov.
\newblock Roberta: A robustly optimized bert pretraining approach.
\newblock \emph{arXiv preprint arXiv:1907.11692}, 2019.

\bibitem[Wang et~al.(2019)Wang, Bi, Yan, Wu, Bao, Xia, Peng, and
  Si]{wang2019structbert}
Wei Wang, Bin Bi, Ming Yan, Chen Wu, Zuyi Bao, Jiangnan Xia, Liwei Peng, and
  Luo Si.
\newblock Structbert: incorporating language structures into pre-training for
  deep language understanding.
\newblock \emph{arXiv preprint arXiv:1908.04577}, 2019.

\bibitem[Vaswani et~al.(2017)Vaswani, Shazeer, Parmar, Uszkoreit, Jones, Gomez,
  Kaiser, and Polosukhin]{vaswani2017attention}
Ashish Vaswani, Noam Shazeer, Niki Parmar, Jakob Uszkoreit, Llion Jones,
  Aidan~N Gomez, {\L}ukasz Kaiser, and Illia Polosukhin.
\newblock Attention is all you need.
\newblock In \emph{Advances in neural information processing systems}, pages
  5998--6008, 2017.

\bibitem[MacQueen et~al.(1967)]{macqueen1967some}
James MacQueen et~al.
\newblock Some methods for classification and analysis of multivariate
  observations.
\newblock In \emph{Proceedings of the fifth Berkeley symposium on mathematical
  statistics and probability}, volume~1, pages 281--297. Oakland, CA, USA,
  1967.

\bibitem[Li et~al.(2021{\natexlab{b}})Li, Bi, Yan, Wang, Huang, Huang, and
  Si]{structurallm}
Chenliang Li, Bin Bi, Ming Yan, Wei Wang, Songfang Huang, Fei Huang, and Luo
  Si.
\newblock Structurallm: Structural pre-training for form understanding.
\newblock \emph{CoRR}, abs/2105.11210, 2021{\natexlab{b}}.
\newblock URL \url{https://arxiv.org/abs/2105.11210}.

\bibitem[Rajpurkar et~al.(2016)Rajpurkar, Zhang, Lopyrev, and
  Liang]{rajpurkar2016squad}
Pranav Rajpurkar, Jian Zhang, Konstantin Lopyrev, and Percy Liang.
\newblock Squad: 100,000+ questions for machine comprehension of text.
\newblock \emph{arXiv preprint arXiv:1606.05250}, 2016.

\bibitem[Chen et~al.(2017)Chen, Fisch, Weston, and Bordes]{chen2017reading}
Danqi Chen, Adam Fisch, Jason Weston, and Antoine Bordes.
\newblock Reading wikipedia to answer open-domain questions.
\newblock \emph{arXiv preprint arXiv:1704.00051}, 2017.

\bibitem[Cai and Vasconcelos(2018)]{cai2018cascade}
Zhaowei Cai and Nuno Vasconcelos.
\newblock Cascade r-cnn: Delving into high quality object detection.
\newblock In \emph{Proceedings of the IEEE conference on computer vision and
  pattern recognition}, pages 6154--6162, 2018.

\bibitem[Pan et~al.(2018)Pan, Luo, Shi, and Tang]{pan2018two}
Xingang Pan, Ping Luo, Jianping Shi, and Xiaoou Tang.
\newblock Two at once: Enhancing learning and generalization capacities via
  ibn-net.
\newblock In \emph{Proceedings of the European Conference on Computer Vision
  (ECCV)}, pages 464--479, 2018.

\bibitem[Hu et~al.(2018)Hu, Shen, and Sun]{hu2018squeeze}
Jie Hu, Li~Shen, and Gang Sun.
\newblock Squeeze-and-excitation networks.
\newblock In \emph{Proceedings of the IEEE conference on computer vision and
  pattern recognition}, pages 7132--7141, 2018.

\bibitem[Jacobs et~al.(1991)Jacobs, Jordan, Nowlan, and
  Hinton]{jacobs1991adaptive}
Robert~A Jacobs, Michael~I Jordan, Steven~J Nowlan, and Geoffrey~E Hinton.
\newblock Adaptive mixtures of local experts.
\newblock \emph{Neural computation}, 3\penalty0 (1):\penalty0 79--87, 1991.

\bibitem[Shazeer et~al.(2017)Shazeer, Mirhoseini, Maziarz, Davis, Le, Hinton,
  and Dean]{shazeer2017outrageously}
Noam Shazeer, Azalia Mirhoseini, Krzysztof Maziarz, Andy Davis, Quoc Le,
  Geoffrey Hinton, and Jeff Dean.
\newblock Outrageously large neural networks: The sparsely-gated
  mixture-of-experts layer.
\newblock \emph{arXiv preprint arXiv:1701.06538}, 2017.

\bibitem[Fedus et~al.(2021)Fedus, Zoph, and Shazeer]{fedus2021switch}
William Fedus, Barret Zoph, and Noam Shazeer.
\newblock Switch transformers: Scaling to trillion parameter models with simple
  and efficient sparsity.
\newblock \emph{arXiv preprint arXiv:2101.03961}, 2021.

\bibitem[Singh et~al.(2019)Singh, Natarajan, Shah, Jiang, Chen, Batra, Parikh,
  and Rohrbach]{textvqa}
Amanpreet Singh, Vivek Natarajan, Meet Shah, Yu~Jiang, Xinlei Chen, Dhruv
  Batra, Devi Parikh, and Marcus Rohrbach.
\newblock Towards {VQA} models that can read.
\newblock \emph{CoRR}, abs/1904.08920, 2019.
\newblock URL \url{http://arxiv.org/abs/1904.08920}.

\bibitem[Biten et~al.(2019)Biten, Tito, Mafla, G{\'{o}}mez, Rusi{\~{n}}ol,
  Valveny, Jawahar, and Karatzas]{stvqa}
Ali~Furkan Biten, Rub{\`{e}}n Tito, Andr{\'{e}}s Mafla, Llu{\'{\i}}s
  G{\'{o}}mez, Mar{\c{c}}al Rusi{\~{n}}ol, Ernest Valveny, C.~V. Jawahar, and
  Dimosthenis Karatzas.
\newblock Scene text visual question answering.
\newblock \emph{CoRR}, abs/1905.13648, 2019.
\newblock URL \url{http://arxiv.org/abs/1905.13648}.

\bibitem[Lin et~al.(2014)Lin, Maire, Belongie, Hays, Perona, Ramanan,
  Doll{\'a}r, and Zitnick]{lin2014microsoft}
Tsung-Yi Lin, Michael Maire, Serge Belongie, James Hays, Pietro Perona, Deva
  Ramanan, Piotr Doll{\'a}r, and C~Lawrence Zitnick.
\newblock Microsoft coco: Common objects in context.
\newblock In \emph{European conference on computer vision}, pages 740--755.
  Springer, 2014.

\bibitem[Antol et~al.(2015)Antol, Agrawal, Lu, Mitchell, Batra,
  Lawrence~Zitnick, and Parikh]{antol2015vqa}
Stanislaw Antol, Aishwarya Agrawal, Jiasen Lu, Margaret Mitchell, Dhruv Batra,
  C~Lawrence~Zitnick, and Devi Parikh.
\newblock Vqa: Visual question answering.
\newblock In \emph{Proceedings of the IEEE international conference on computer
  vision}, pages 2425--2433, 2015.

\bibitem[Hudson and Manning(2019)]{hudson2019gqa}
Drew~A Hudson and Christopher~D Manning.
\newblock Gqa: A new dataset for real-world visual reasoning and compositional
  question answering.
\newblock In \emph{Proceedings of the IEEE Conference on Computer Vision and
  Pattern Recognition}, pages 6700--6709, 2019.

\bibitem[Zhu et~al.(2016)Zhu, Groth, Bernstein, and Fei-Fei]{zhu2016visual7w}
Yuke Zhu, Oliver Groth, Michael Bernstein, and Li~Fei-Fei.
\newblock Visual7w: Grounded question answering in images.
\newblock In \emph{Proceedings of the IEEE conference on computer vision and
  pattern recognition}, pages 4995--5004, 2016.

\bibitem[Sharma et~al.(2018)Sharma, Ding, Goodman, and
  Soricut]{sharma2018conceptual}
Piyush Sharma, Nan Ding, Sebastian Goodman, and Radu Soricut.
\newblock Conceptual captions: A cleaned, hypernymed, image alt-text dataset
  for automatic image captioning.
\newblock In \emph{Proceedings of the 56th Annual Meeting of the Association
  for Computational Linguistics (Volume 1: Long Papers)}, pages 2556--2565,
  2018.

\bibitem[Lewis et~al.(2006)Lewis, Agam, Argamon, Frieder, Grossman, and
  Heard]{Lewis}
D.~Lewis, G.~Agam, S.~Argamon, O.~Frieder, D.~Grossman, and J.~Heard.
\newblock Building a test collection for complex document information
  processing.
\newblock SIGIR '06, page 665–666, New York, NY, USA, 2006. Association for
  Computing Machinery.
\newblock ISBN 1595933697.
\newblock \doi{10.1145/1148170.1148307}.
\newblock URL \url{https://doi.org/10.1145/1148170.1148307}.

\bibitem[Xie et~al.(2017)Xie, Girshick, Doll{\'a}r, Tu, and
  He]{xie2017aggregated}
Saining Xie, Ross Girshick, Piotr Doll{\'a}r, Zhuowen Tu, and Kaiming He.
\newblock Aggregated residual transformations for deep neural networks.
\newblock In \emph{Proceedings of the IEEE conference on computer vision and
  pattern recognition}, pages 1492--1500, 2017.

\bibitem[Yu et~al.(2019{\natexlab{a}})Yu, Yu, Cui, Tao, and Tian]{yu2019mcan}
Zhou Yu, Jun Yu, Yuhao Cui, Dacheng Tao, and Qi~Tian.
\newblock Deep modular co-attention networks for visual question answering.
\newblock In \emph{Proceedings of the IEEE Conference on Computer Vision and
  Pattern Recognition (CVPR)}, pages 6281--6290, 2019{\natexlab{a}}.

\bibitem[Gan et~al.(2020)Gan, Chen, Li, Zhu, Cheng, and Liu]{gan2020large}
Zhe Gan, Yen-Chun Chen, Linjie Li, Chen Zhu, Yu~Cheng, and Jingjing Liu.
\newblock Large-scale adversarial training for vision-and-language
  representation learning.
\newblock In \emph{NeurIPS}, 2020.

\bibitem[Guo et~al.(2019)Guo, Xu, and Tao]{guo2019bilinear}
Dalu Guo, Chang Xu, and Dacheng Tao.
\newblock Bilinear graph networks for visual question answering.
\newblock \emph{arXiv preprint arXiv:1907.09815}, 2019.

\bibitem[Lin et~al.(2020)Lin, Yang, Zhang, Liu, Zhou, and
  Yang]{lin2020interbert}
Junyang Lin, An~Yang, Yichang Zhang, Jie Liu, Jingren Zhou, and Hongxia Yang.
\newblock Interbert: Vision-and-language interaction for multi-modal
  pretraining.
\newblock \emph{arXiv preprint arXiv:2003.13198}, 2020.

\bibitem[Cui et~al.(2021)Cui, Yu, Wang, Zhao, Zhang, Wang, and
  Yu]{cui2021rosita}
Yuhao Cui, Zhou Yu, Chunqi Wang, Zhongzhou Zhao, Ji~Zhang, Meng Wang, and Jun
  Yu.
\newblock Rosita: Enhancing vision-and-language semantic alignments via cross-
  and intra-modal knowledge integration, 2021.

\bibitem[Li et~al.(2020{\natexlab{a}})Li, Gao, Niu, Xiao, Liu, Liu, Wu, and
  Wang]{li2020unimo}
Wei Li, Can Gao, Guocheng Niu, Xinyan Xiao, Hao Liu, Jiachen Liu, Hua Wu, and
  Haifeng Wang.
\newblock Unimo: Towards unified-modal understanding and generation via
  cross-modal contrastive learning.
\newblock \emph{arXiv preprint arXiv:2012.15409}, 2020{\natexlab{a}}.

\bibitem[Wang et~al.(2021)Wang, Yu, Yu, Dai, Tsvetkov, and Cao]{wang2021simvlm}
Zirui Wang, Jiahui Yu, Adams~Wei Yu, Zihang Dai, Yulia Tsvetkov, and Yuan Cao.
\newblock Simvlm: Simple visual language model pretraining with weak
  supervision.
\newblock \emph{CoRR}, abs/2108.10904, 2021.

\bibitem[Li et~al.(2020{\natexlab{b}})Li, Yin, Li, Zhang, Hu, Zhang, Wang, Hu,
  Dong, Wei, et~al.]{li2020oscar}
Xiujun Li, Xi~Yin, Chunyuan Li, Pengchuan Zhang, Xiaowei Hu, Lei Zhang, Lijuan
  Wang, Houdong Hu, Li~Dong, Furu Wei, et~al.
\newblock Oscar: Object-semantics aligned pre-training for vision-language
  tasks.
\newblock In \emph{European Conference on Computer Vision}, pages 121--137.
  Springer, 2020{\natexlab{b}}.

\bibitem[Lu et~al.(2019)Lu, Batra, Parikh, and Lee]{lu2019vilbert}
Jiasen Lu, Dhruv Batra, Devi Parikh, and Stefan Lee.
\newblock Vilbert: Pretraining task-agnostic visiolinguistic representations
  for vision-and-language tasks.
\newblock In \emph{Advances in Neural Information Processing Systems}, pages
  13--23, 2019.

\bibitem[Lu et~al.(2020)Lu, Goswami, Rohrbach, Parikh, and Lee]{Lu_2020_CVPR}
Jiasen Lu, Vedanuj Goswami, Marcus Rohrbach, Devi Parikh, and Stefan Lee.
\newblock 12-in-1: Multi-task vision and language representation learning.
\newblock In \emph{The IEEE/CVF Conference on Computer Vision and Pattern
  Recognition (CVPR)}, June 2020.

\bibitem[Fukui et~al.(2016)Fukui, Park, Yang, Rohrbach, Darrell, and
  Rohrbach]{fukui2016multimodal}
Akira Fukui, Dong~Huk Park, Daylen Yang, Anna Rohrbach, Trevor Darrell, and
  Marcus Rohrbach.
\newblock Multimodal compact bilinear pooling for visual question answering and
  visual grounding.
\newblock \emph{arXiv preprint arXiv:1606.01847}, 2016.

\bibitem[Yu et~al.(2017)Yu, Yu, Fan, and Tao]{yu2017multi}
Zhou Yu, Jun Yu, Jianping Fan, and Dacheng Tao.
\newblock Multi-modal factorized bilinear pooling with co-attention learning
  for visual question answering.
\newblock In \emph{Proceedings of the IEEE international conference on computer
  vision}, pages 1821--1830, 2017.

\bibitem[Lu et~al.(2016)Lu, Yang, Batra, and Parikh]{lu2016hierarchical}
Jiasen Lu, Jianwei Yang, Dhruv Batra, and Devi Parikh.
\newblock Hierarchical question-image co-attention for visual question
  answering.
\newblock \emph{Advances in neural information processing systems},
  29:\penalty0 289--297, 2016.

\bibitem[Yang et~al.(2016)Yang, He, Gao, Deng, and Smola]{yang2016stacked}
Zichao Yang, Xiaodong He, Jianfeng Gao, Li~Deng, and Alex Smola.
\newblock Stacked attention networks for image question answering.
\newblock In \emph{Proceedings of the IEEE conference on computer vision and
  pattern recognition}, pages 21--29, 2016.

\bibitem[Yu et~al.(2019{\natexlab{b}})Yu, Yu, Cui, Tao, and Tian]{yu2019deep}
Zhou Yu, Jun Yu, Yuhao Cui, Dacheng Tao, and Qi~Tian.
\newblock Deep modular co-attention networks for visual question answering.
\newblock In \emph{Proceedings of the IEEE/CVF Conference on Computer Vision
  and Pattern Recognition}, pages 6281--6290, 2019{\natexlab{b}}.

\bibitem[Li et~al.(2020{\natexlab{c}})Li, Duan, Fang, Gong, and
  Jiang]{li2020unicoder}
Gen Li, Nan Duan, Yuejian Fang, Ming Gong, and Daxin Jiang.
\newblock Unicoder-vl: A universal encoder for vision and language by
  cross-modal pre-training.
\newblock In \emph{Proceedings of the AAAI Conference on Artificial
  Intelligence}, volume~34, pages 11336--11344, 2020{\natexlab{c}}.

\bibitem[Nguyen et~al.(2020)Nguyen, Goswami, and Chen]{nguyen2020movie}
Duy-Kien Nguyen, Vedanuj Goswami, and Xinlei Chen.
\newblock Movie: Revisiting modulated convolutions for visual counting and
  beyond.
\newblock \emph{arXiv preprint arXiv:2004.11883}, 2020.

\bibitem[Wu et~al.(2016)Wu, Wang, Shen, Dick, and Van Den~Hengel]{wu2016ask}
Qi~Wu, Peng Wang, Chunhua Shen, Anthony Dick, and Anton Van Den~Hengel.
\newblock Ask me anything: Free-form visual question answering based on
  knowledge from external sources.
\newblock In \emph{Proceedings of the IEEE conference on computer vision and
  pattern recognition}, pages 4622--4630, 2016.

\bibitem[Chen et~al.(2015)Chen, Fang, Lin, Vedantam, Gupta, Doll{\'a}r, and
  Zitnick]{chen2015microsoft}
Xinlei Chen, Hao Fang, Tsung-Yi Lin, Ramakrishna Vedantam, Saurabh Gupta, Piotr
  Doll{\'a}r, and C~Lawrence Zitnick.
\newblock Microsoft coco captions: Data collection and evaluation server.
\newblock \emph{arXiv preprint arXiv:1504.00325}, 2015.

\bibitem[Kuznetsova et~al.(2020)Kuznetsova, Rom, Alldrin, Uijlings, Krasin,
  Pont-Tuset, Kamali, Popov, Malloci, Kolesnikov, et~al.]{kuznetsova2020open}
Alina Kuznetsova, Hassan Rom, Neil Alldrin, Jasper Uijlings, Ivan Krasin, Jordi
  Pont-Tuset, Shahab Kamali, Stefan Popov, Matteo Malloci, Alexander
  Kolesnikov, et~al.
\newblock The open images dataset v4.
\newblock \emph{International Journal of Computer Vision}, 128\penalty0
  (7):\penalty0 1956--1981, 2020.

\bibitem[Shao et~al.(2019)Shao, Li, Zhang, Peng, Yu, Zhang, Li, and
  Sun]{shao2019objects365}
Shuai Shao, Zeming Li, Tianyuan Zhang, Chao Peng, Gang Yu, Xiangyu Zhang, Jing
  Li, and Jian Sun.
\newblock Objects365: A large-scale, high-quality dataset for object detection.
\newblock In \emph{Proceedings of the IEEE/CVF International Conference on
  Computer Vision}, pages 8430--8439, 2019.

\bibitem[Carion et~al.(2020)Carion, Massa, Synnaeve, Usunier, Kirillov, and
  Zagoruyko]{carion2020end}
Nicolas Carion, Francisco Massa, Gabriel Synnaeve, Nicolas Usunier, Alexander
  Kirillov, and Sergey Zagoruyko.
\newblock End-to-end object detection with transformers.
\newblock In \emph{European Conference on Computer Vision}, pages 213--229.
  Springer, 2020.

\bibitem[Suhr et~al.(2018)Suhr, Zhou, Zhang, Zhang, Bai, and
  Artzi]{suhr2018corpus}
Alane Suhr, Stephanie Zhou, Ally Zhang, Iris Zhang, Huajun Bai, and Yoav Artzi.
\newblock A corpus for reasoning about natural language grounded in
  photographs.
\newblock \emph{arXiv preprint arXiv:1811.00491}, 2018.

\end{thebibliography}
